\documentclass[10pt,journal,compsoc]{IEEEtran}
\ifCLASSOPTIONcompsoc
  \usepackage[nocompress]{cite}
\else
  \usepackage{cite}
\fi
\usepackage{xcolor}
\usepackage{amsmath}
\usepackage{amssymb}
\usepackage{booktabs}
\usepackage{algorithm}
\usepackage{multirow}
\usepackage{algorithmic}
\usepackage{graphicx}
\usepackage{bm}
\usepackage{color}
\usepackage{colortbl}
\usepackage{url}
\usepackage{float}
\usepackage{subcaption}
\usepackage{indentfirst}
\usepackage{ragged2e}
\usepackage{verbatim}

\usepackage[pagebackref,breaklinks,colorlinks]{hyperref}
\usepackage{makecell}
\usepackage{xspace}
\usepackage{pifont}
\newcommand{\cmark}{\ding{51}}  
\newcommand{\xmark}{\ding{55}}  
\usepackage{lineno} 
\usepackage{stfloats}

\definecolor{lightgray}{gray}{0.9}

\makeatletter
\DeclareRobustCommand\onedot{\futurelet\@let@token\@onedot}
\def\@onedot{\ifx\@let@token.\else.\null\fi\xspace}

\def\eg{\emph{e.g}\onedot} 
\def\ie{\emph{i.e}\onedot}

\makeatother

\ifCLASSINFOpdf
\else
\fi

\begin{document}
\title{SAPNet++: Evolving Point-Prompted Instance Segmentation with Semantic and Spatial Awareness}

\author{Zhaoyang Wei, Xumeng Han, Xuehui Yu, Xue Yang,~\IEEEmembership{Member,~IEEE}, Guorong Li,~\IEEEmembership{Senior Member,~IEEE},  Zhenjun Han,~\IEEEmembership{Member,~IEEE}, Jianbin Jiao,~\IEEEmembership{Member,~IEEE} 


\IEEEcompsocitemizethanks{\IEEEcompsocthanksitem Corresponding author: Zhenjun Han.}
\IEEEcompsocitemizethanks{\IEEEcompsocthanksitem Z. Wei, X. Han, X. Yu, Z. Han and J. Jiao are with the School of Electronic, Electrical and Communication Engineering, University of Chinese Academy of Science (UCAS), Beijing, 100049, China. E-mail: \{weizhaoyang23, hanxumeng19, yuxuehui17\}@mails.ucas.ac.cn, \{hanzhj, jiaojb\}@ucas.ac.cn.}
\IEEEcompsocitemizethanks{\IEEEcompsocthanksitem X. Yang is with School of Automation and Intelligent Sensing, Shanghai Jiao Tong University (SJTU), Shanghai, 200240, China. E-mail: yangxue-2019-sjtu@sjtu.edu.cn}
\IEEEcompsocitemizethanks{\IEEEcompsocthanksitem G. Li is with the School of Computer Science and Technology, University of Chinese Academy of Science (UCAS), Beijing, 100049, China. E-mail: liguorong@ucas.ac.cn.}}


\markboth{Journal of \LaTeX\ Class Files,~Vol.~14, No.~8, August~2015}%
{Shell \MakeLowercase{\textit{et al.}}: }

\IEEEtitleabstractindextext{%
\justify
\begin{abstract}
\linenumbers
Single-point annotation is increasingly prominent in visual tasks for labeling cost reduction. However, it challenges tasks requiring high precision, such as the point-prompted instance segmentation (PPIS) task, which aims to estimate precise masks using single-point prompts to train a segmentation network. Due to the constraints of point annotations, \textbf{granularity ambiguity} and \textbf{boundary uncertainty} arise \ie{, the difficulty distinguishing between different levels of detail (\eg{, whole object vs. parts}) and the challenge of precisely delineating object boundaries.}
Previous works have usually inherited the paradigm of mask generation along with proposal selection to achieve PPIS.
However, proposal selection relies solely on category information, failing to resolve the ambiguity of different granularity. Furthermore, mask generators offer only finite discrete solutions that often deviate from actual masks, particularly at boundaries.
To address these issues, we propose the Semantic-Aware Point-Prompted Instance Segmentation Network (SAPNet). It integrates Point Distance Guidance and Box Mining Strategy to tackle group and local issues caused by the point's granularity ambiguity. Additionally, we incorporate completeness scores within proposals to add spatial granularity awareness, enhancing multiple instance learning (MIL) in proposal selection termed S-MIL. The Multi-level Affinity Refinement conveys pixel and semantic clues, narrowing boundary uncertainty during mask refinement. These modules culminate in SAPNet++, mitigating point prompt's granularity ambiguity and boundary uncertainty and significantly improving segmentation performance.
Extensive experiments on four challenging datasets validate the effectiveness of our methods, highlighting the potential to advance PPIS.

\end{abstract}

\begin{IEEEkeywords}
Instance Segmentation, Point Prompt, Granularity Ambiguity, Boundary Uncertainty.
\end{IEEEkeywords}}

\maketitle

\IEEEdisplaynontitleabstractindextext

\IEEEpeerreviewmaketitle

\IEEEraisesectionheading{\section{Introduction}\label{sec:introduction}}

\IEEEPARstart{I}nstance segmentation, a pivotal task in computer vision, aims to accurately discern pixel-level labels for instances of interest and their semantic content within images. This capability is crucial across various domains, including autonomous driving, image editing, and human-computer interaction. High-performing methods, like Mask R-CNN~\cite{DBLP:MASK-RCNN}, SOLOv2~\cite{DBLP:solov2}, Mask2Former~\cite{DBLP:mask2former}, and others~\cite{DBLP:SOLO,DBLP:condinst,DBLP:yolact}, have shown impressive results but require costly pixel-wise labeling, which is a significant bottleneck, especially for large datasets such as Cityscapes, where labeling an image can take over 1.5 hours on average~\cite{DBLP:cityscapes}.

To mitigate the reliance on costly and laborious pixel-level annotations, weakly supervised instance segmentation (WSIS) has garnered significant attention in recent research. These approaches leverage more cost-effective forms of supervision, such as bounding boxes~\cite{DBLP:boxlevelset,DBLP:Boxinst,DBLP:discobox,yang2023h2rbox,yu2023h2rboxv2}, sparse points~\cite{DBLP:point2mask,DBLP:PSPS}, or image-level labels~\cite{DBLP:BESTIE,DBLP:IRNet}. Consequently, WSIS lowers the deployment barrier for advanced computer vision models in practical applications and has potential to align with the trend towards reducing annotation granularity while preserving performance.

\textcolor{black}{However, despite their promise, accurately delineating instance boundaries remains a common challenge for weakly supervised methods. Many existing image-level methods~\cite{DBLP:wsddn, DBLP:BESTIE} still struggle to directly obtain high-quality segmentation results, often needing to rely on iterative refinement processes or the generation of pseudo-labels to serve as supervision for model training, which adds complexity to the pipeline. 
Despite providing richer object extent information and yielding promising results without iterative refinement, box-supervised methods~\cite{DBLP:discobox,DBLP:box2mask,DBLP:boxlevelset,DBLP:Boxinst} are often constrained by relatively high annotation costs and long training times. 
Offering a more economical alternative to box-level annotations and richer spatial cues than image-level labels, point-supervised methods~\cite{DBLP:BESTIE, DBLP:PSPS, DBLP:point2mask} still exhibit a significant performance disparity compared to their fully/box-supervised counterparts. This performance gap highlights the need for more powerful frameworks that can better leverage sparse annotations.}

\begin{figure*}[htb]
\begin{center}
    \begin{tabular}{ccc}
    \includegraphics[width=0.9\linewidth]{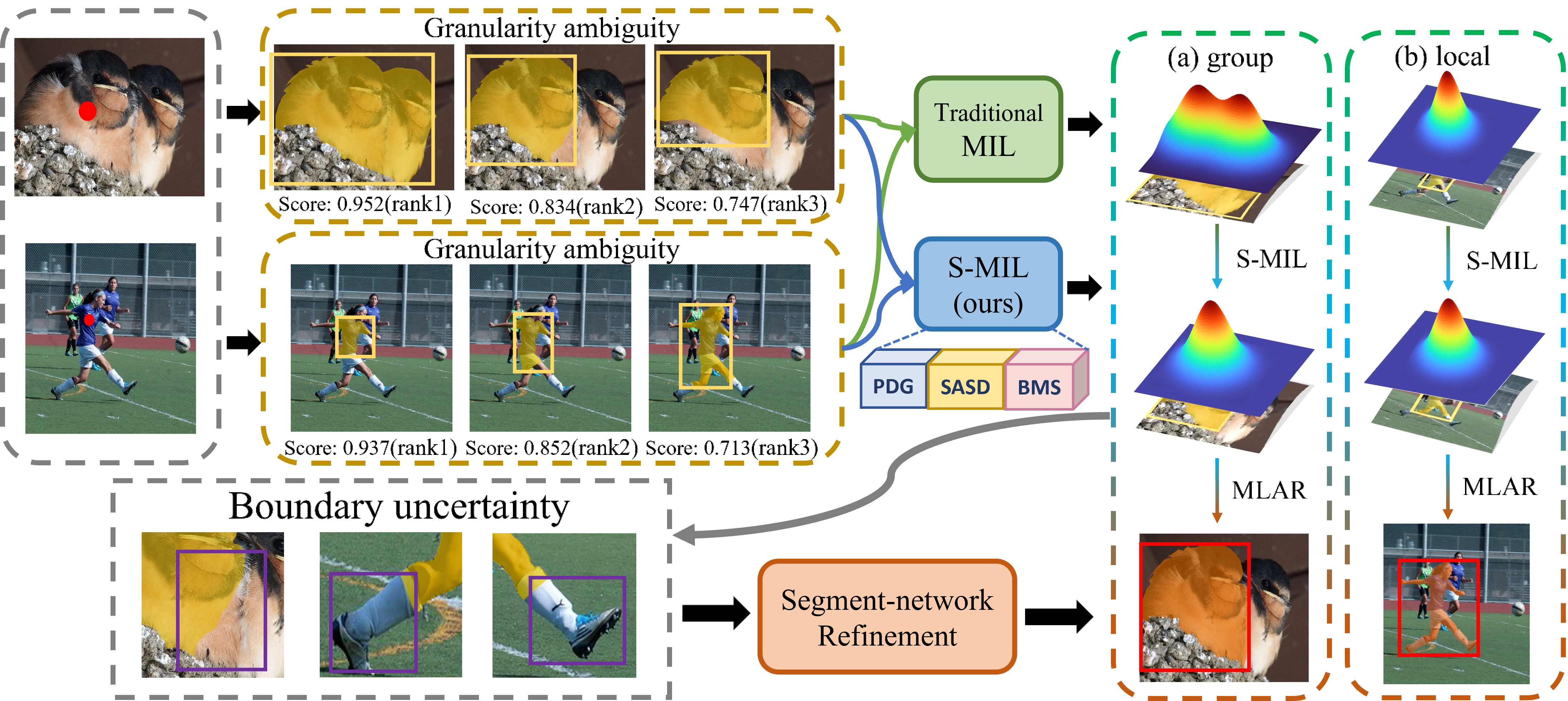}
    \end{tabular}
    \caption{The motivation behind SAPNet++ stems from two key challenges of point annotation:
\textbf{(a) Granularity Ambiguity}: SAM-generated masks often assign higher scores to non-target categories (\eg{, the clothes instead of the person in the yellow dashed box}). Additionally, segmentation struggles to separate individual targets within the same category (group issue) and overlooks local details due to MIL's preference under point labels for foreground-dominant regions (local issue) in the green dashed box. Within the blue dashed box,our proposed S-MIL tackles these issues by selecting proposals that capture the target's complete semantic information.
\textbf{(b) Boundary Uncertainty}: Despite resolving granularity ambiguity and related issues, compared to the ground truth, most proposal generators still yield proposals of varying quality, leading to ``boundary uncertainty'' where masks do not fully encompass the targets. We address that by refining predicted masks during segmentation to achieve highly satisfactory results even with imprecise supervision.
}

\label{fig:issues}
\end{center}
\vspace{-10pt}
\end{figure*} 
To bridge the performance gap, we define the paradigm of \textbf{Point-Prompted Instance Segmentation (PPIS)}, which aims to shift the role of a point from a mere supervisory signal to a continuous guide throughout the segmentation process. The advent of powerful foundation models, particularly the Segment Anything Model (SAM)~\cite{DBLP:SAM}, makes this paradigm highly feasible. SAM is renowned for its formidable zero-shot capabilities and, crucially, its promptable architecture that can accept diverse inputs, including point prompts, to generate corresponding masks. While SAM's zero-shot segmentation capability is formidable, its direct application in PPIS gives rise to two critical challenges, stemming from the interplay between point sparsity and SAM's intrinsic properties:\\
\textbf{1)} \textbf{Granularity Ambiguity:} This issue arises from the confluence of a point's spatial sparsity and SAM's design to produce masks at multiple granularities. A single point provides insufficient information to disambiguate the correct instance scope, often causing the model to isolate only a part of the target object rather than its entirety (\eg, segmenting only ``clothing'' or ``upper body'' when a person is clicked, as shown in Fig.~\ref{fig:issues}).\\
\textbf{2)} \textbf{Boundary Uncertainty:} This is exacerbated by SAM's class-agnostic nature. Lacking semantic guidance, SAM may erroneously merge a target with adjacent but distinct objects into a single mask (\eg, including a neighboring bird, as shown in Fig.~\ref{fig:issues}), resulting in uncertain boundaries.

To address the aforementioned challenges, we first integrate the Multiple Instance Learning (MIL)~\cite{DBLP:MIL} as a selection mechanism with SAM. Because, the MIL is built upon the principle of grouping proposals into bags, where it is assumed that every positive bag contains at least one true instance. However, conventional MIL's sole reliance on classification scores renders it spatially agnostic. This leads to persistent ``local'' and ``group'' issues (as shown in Fig.~\ref{fig:issues}), meaning that \textbf{granularity ambiguity} is not fully resolved.

To overcome this limitation, we first enhance the conventional Multi-Instance Learning (MIL) framework in SAPNet. Traditional MIL is spatially agnostic, as it relies solely on classification scores, often leading to the erroneous selection of proposals that either cover only the most discriminative part of an object (``local'' issue) or incorrectly merge adjacent instances (``group'' issue). SAPNet mitigates these problems by incorporating \textbf{point-distance guidance}, which penalizes proposals encompassing multiple annotated instances to resolve the ``group'' issue, and an \textbf{adaptive box mining} strategy, which encourages the selection of more spatially complete proposals to alleviate the ``local'' issue. These mechanisms enable dynamic re-scoring and initial refinement of proposals, substantially reducing granularity errors.

However, to endow the model with genuine spatial awareness, SAPNet++ introduces a more fundamental solution: \textbf{Spatial-Aware Self-Distillation (SASD)}, which forms our S-MIL. The core innovation of SASD is the concept of \textbf{``completeness''}, where the model is explicitly trained to predict the spatial integrity of the target within each proposal. By supervising this completeness score, S-MIL transcends the limitations of classification-based selection. It learns to identify proposals that are not only semantically correct but also spatially complete, effectively distinguishing between partial and whole object representations. As depicted in the blue dashed box in Fig.~\ref{fig:issues}, S-MIL selects proposals that best capture the target's complete semantic and spatial extent, representing a significant enhancement over traditional MIL by resolving the granularity ambiguity.

Finally, to tackle the \textbf{boundary uncertainty} problem, we recognize that even the best-selected proposals from S-MIL may yield masks with coarse or imprecise contours. Directly supervising a segmentation network with these imperfect pseudo-masks would inherently limit its performance. Therefore, SAPNet++ integrates a \textbf{Multi-level Affinity Refinement (MLAR)} module to refine these masks. This mechanism operates by propagating affinity relationships across multiple scales. Specifically, it merges global affinity, which captures long-range contextual dependencies, with cascaded local affinity, which focuses on fine-grained details, across both low-dimensional pixel (e.g., color, texture) and high-dimensional semantic spaces. This process effectively polishes the coarse masks, filling in holes and sharpening contours to generate high-quality soft pseudo-labels. These refined labels, in conjunction with the masks chosen by S-MIL, provide robust and precise supervision, enabling the segmentation network to learn accurate object boundaries in a fully end-to-end process.

The contributions of this paper are as follows:

1) We introduce a novel end-to-end paradigm for Point-Prompted Instance Segmentation (PPIS) and systematically investigate its core challenges, namely granularity ambiguity and boundary uncertainty.

2) We propose SAPNet, which resolves the granularity ambiguity inherent in point annotations. It enhances MIL-based proposal selection with point distance guidance and a box mining strategy to effectively mitigate the ``local'' and ``group'' selection challenges.

3) We further develop SAPNet++, which introduces two novel components. A \textbf{Spatial-Aware Self-Distillation (SASD)} module resolves residual granularity ambiguity by explicitly modeling proposal completeness. Concurrently, a \textbf{Multi-level Affinity Refinement (MLAR)} module tackles boundary uncertainty, enabling the generation of high-quality segmentation masks from imprecise supervision.

4) SAPNet++, achieves state-of-the-art (SOTA) performance on multiple challenging benchmarks for PPIS, significantly narrowing the performance gap between point-prompted and fully-supervised methods.

\textit{This article builds on our \textbf{CVPR'24 highlight} conference paper~\cite{DBLP:SAPNet} with significant extensions:
1) We conduct additional ablation experiments as is shown in Tab.\ref{tab:gap} and define specific metrics in Sec.~\ref{sec:analysis} to assess the effectiveness of our approach in addressing MIL's ``group'' and ``local'' issues under weak annotations.
2) We remove the impractical Multi-mask Proposals Supervision (MPS) from SAPNet and refine the training loss function, discussed in detail in Sec.~\ref{sec:loss}.
3) Sec.~\ref{sec:method} outlines the structure of SAPNet++, which uses semantic selection and enhancement branches to generate and refine pseudo labels for supervising the segmentation branch. The Spatial-aware Self-distillation, which assesses target completeness within proposals, is detailed in Sec.~\ref{sec: SASD}.
4) The Multi-level Affinity Refinement (MLAR), enhancing segmentation quality through global and cascaded local affinity, is elaborated in Sec.~\ref{sec:affinity}.
5) Detailed ablation studies of SAPNet++ are presented in Sec.~\ref{sec:ablation study}.
6) Experimental validation and visual analysis of SAPNet and SAPNet++ across multiple datasets, are documented in Sec.~\ref{sec:experiment}, demonstrating their adaptability and effectiveness.}

\section{Related Work}
\textbf{Weakly-Supervised Instance Segmentation.}
Weakly-Supervised Instance Segmentation (WSIS) is a key paradigm for generating precise object masks with minimal supervision, adapting to annotation levels from image labels to bounding boxes. Recent WSIS research aims to close the performance gap between weakly- and fully-supervised methods, with a focus on approaches using box-level \cite{DBLP:Boxinst,DBLP:BBTP} and image-level annotations \cite{DBLP:BESTIE,DBLP:inter-pixel}.
Box-based methods like BBTP \cite{DBLP:BBTP} and BoxInst \cite{DBLP:Boxinst} improve accuracy with structural constraints, for instance by treating segmentation as a multiple-instance learning problem or enforcing feature consistency (e.g., color) within frameworks like CondInst \cite{DBLP:condinst}. However, these approaches can increase training complexity and often focus on local features and proposal generation (e.g., using MCG \cite{DBLP:MCG} or SS \cite{DBLP:conf/iccv/SandeUGS11}), thus neglecting the global object shape. In contrast, proposal-free strategies like IRNet \cite{DBLP:IRNet} use class relationships to build masks but can struggle with precise instance delineation. To improve object integrity, methods such as Discobox \cite{DBLP:discobox} and BESTIE \cite{DBLP:BESTIE} incorporate semantic insights, using techniques like pairwise losses or saliency cues \cite{DBLP:filament,DBLP:Boxinst, DBLP:solov2}. Despite these advances, challenges like semantic drift persist; labeling inaccuracies or detection failures can yield poor pseudo-labels \cite{DBLP:consistent,DBLP:mining}, degrading final segmentation quality.

\textbf{Pointly-Supervised Detection and Segmentation.}
Pointly-supervised methods balance minimal annotation cost with high localization accuracy. Methods like WISE-Net \cite{DBLP:WISE-Net}, P2BNet \cite{DBLP:cpf}, and BESTIE \cite{DBLP:BESTIE} use point annotations to overcome the vague localizations common in other weakly-supervised approaches. Similarly, oriented object detection methods like the PointOBB \cite{luo2024pointobb,ren2025pointobbv2,zhang2025pointobbv3} and Point2RBox \cite{yu2024point2rbox,yu2025point2rboxv2} series use point supervision to improve detection of rotated or complex objects. Compared to image-level annotation, the point-based approach has a comparable speed with a modest cost increase (~10\%), yet is significantly faster than box or mask annotation. Crucially, it also mitigates semantic bias, a known challenge in weakly-supervised settings \cite{DBLP:semantic_bias}.
In point-supervised instance segmentation, methods like WISE-Net~\cite{DBLP:WISE-Net}, Point2Mask \cite{DBLP:point2mask}, and PSPS \cite{DBLP:PSPS} demonstrate that a single point can generate precise mask proposals. For example, WISE-Net localizes objects to select suitable masks, while BESTIE uses instance cues and self-correction to reduce semantic drift. Point annotations also serve in refinement; PointRend \cite{DBLP:pointrend}, for instance, uses multiple points to enhance localization accuracy. Furthermore, Attnshift \cite{DBLP:Attnshift} reconstructs entire objects from single-point annotations, extending simple supervision to complex segmentation tasks. Despite these advances, the efficacy of point-supervised methods in complex scenarios remains an open research area. Studies by Li \textit{et al.} \cite{DBLP:segmentation_challenge} and Zhou \textit{et al.} \cite{DBLP:zhou_efficiency} offer key insights into their scalability and real-world performance, guiding future developments.

\textbf{Prompt Learning and Foundation Models.}
Prompt-based learning with foundation models is a key advance towards adaptable, scalable, and efficient AI systems. Trained on diverse, extensive datasets, these models exhibit strong zero-shot generalization capabilities \cite{DBLP:learning-transfer, DBLP:scaling, DBLP:cnn}. A prime example in computer vision is SAM, which uses an innovative data engine for model-in-the-loop dataset annotation \cite{DBLP:SAM}.
Prompting, a shift from conventional fine-tuning, allows foundation models like SAM to adapt to new tasks without extensive retraining \cite{DBLP:ovarnet, DBLP:power, DBLP:promptdet, DBLP:visual}. This improves SAM's real-world applicability, especially with scarce or incomplete training data. Building on this, SAM-based adaptations like Fast-SAM, HQ-SAM, and Semantic-SAM optimize for speed, segmentation quality, and semantic accuracy, respectively \cite{DBLP:fast-sam, DBLP:HQ-sam, DBLP:semantic-sam}. SAM's adaptability is further shown by its use in specialized fields like remote sensing with Rsprompter \cite{DBLP:rsprompter}, medical imaging, and video tracking, often using SAM-derived pseudo-labels \cite{DBLP:a-sam,DBLP:sam-wsss1,DBLP:sam-wsss2,han2025boosting}. Our research builds on these innovations, using point annotations to generate mask proposals for instance segmentation and significantly enhance performance.

\textbf{Affinity in Weakly-supervised Image Segmentation.}
In weakly-supervised image segmentation, modeling pixel affinity is pivotal for capturing semantic relationships and enhancing accuracy. Traditional methods like Conditional Random Fields (CRFs) \cite{DBLP:crf} use color and spatial data to model pairwise relationships, often as a network layer or a post-processing step \cite{DBLP:recurrentcrf,DBLP:gatedcrf}.
Recent methods like BoxInst \cite{DBLP:Boxinst} and DiscoBox \cite{DBLP:discobox} use the LAB color space and Gaussian kernels to model local pixel affinity. Moreover, modeling semantic affinity between pixels further enhances segmentation quality. Some weakly-supervised methods \cite{DBLP:inter-pixel,DBLP:affinitynet} refine pseudo-labels using pairwise affinity on semantic maps to generate high-quality supervision.
While effective for local relationships, these methods often neglect global context and long-range dependencies, thus failing to capture the full object topology \cite{DBLP:learnabletree,DBLP:shortest}. To address this, graph-based and attention-based methods have been proposed to capture global topological information \cite{DBLP:attentionaffinity,DBLP:treeloss}. For instance, AFA \cite{DBLP:afa} integrates affinity with transformers, leveraging their global information capacity to improve consistency between self-attention and semantic affinity for better initial pseudo-labels in Weakly-Supervised Semantic Segmentation. Similarly, APro \cite{DBLP:apro} uses minimum spanning trees to capture object topology and long-range dependencies, constructing high-quality pseudo-labels from global affinity.
\section{Methodology}\label{sec:method}
\subsection{Overview}
\textcolor{black}{Acquiring the precise pixel-level annotations required for accurate instance segmentation is often prohibitively costly and labor-intensive, presenting a significant bottleneck.
To tackle this challenge, we present SAPNet, a novel network that leverages single-point prompts to achieve precise, category-specific instance segmentation with substantially reduced manual labeling overhead. }

\begin{figure*}[tb!]
\begin{center}
    \includegraphics[width=0.9\linewidth]{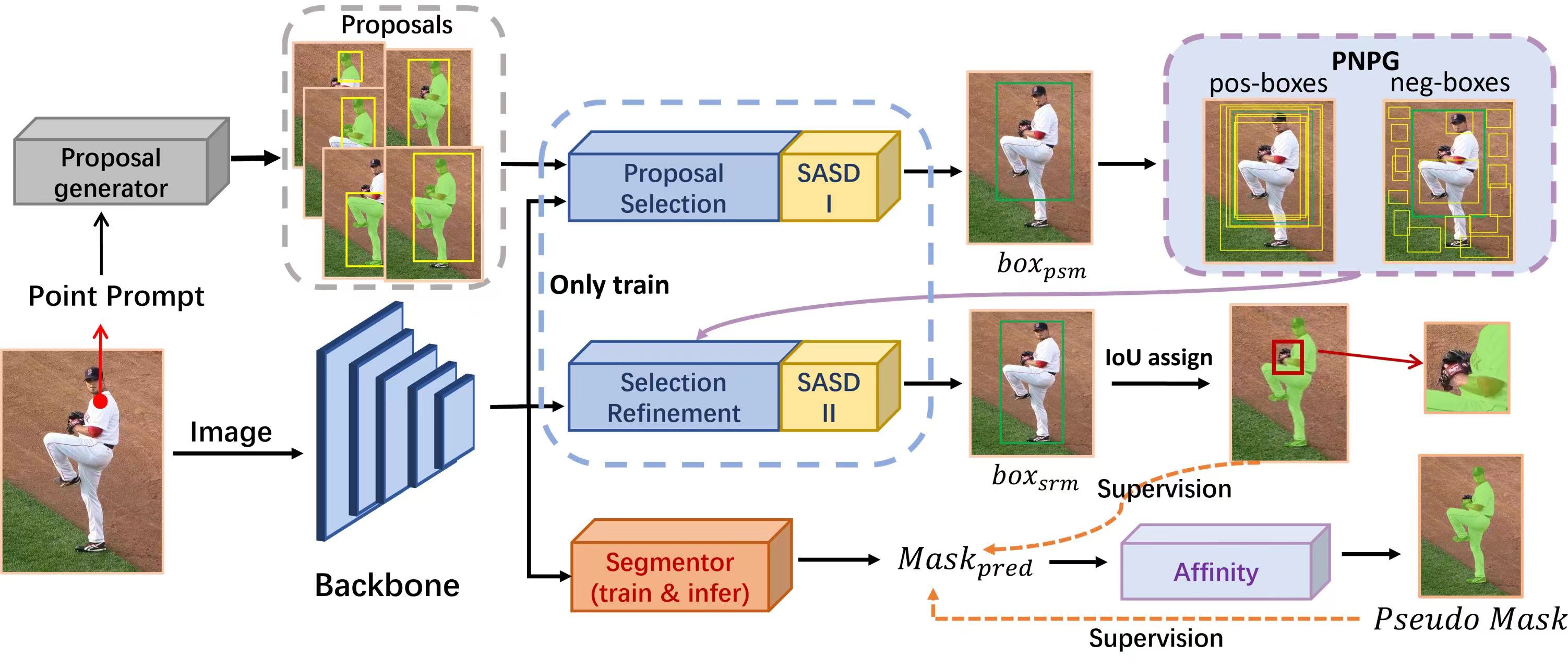}
    \vspace{-5pt}
    \caption{
    SAPNet++ is structured into three key branches: the proposal selection mechanism branch for the initial proposal selection using proposal selection and SASD I, the selection refinement mechanism branch for refining proposals through SASD II and semantic matching, and the SEG branch for segmentation and further refinement. \textbf{i). Proposal Selection Mechanism (PSM) Branch:} The process starts with generating category-agnostic mask proposals using point prompts within a visual foundation model. The PSM branch employs multi-instance learning and a point-guided strategy to construct the PSM for the input box proposals. It combines completeness scores from SASD I with confidence scores to capture global object semantics, initially filtering box proposals to obtain the \(box_{psm}\). \textbf{ii). Selection Refinement Mechanism (SRM) Branch:} Emulating typical MIL approaches, this branch utilizes positive and negative proposal generator to create high-quality positive and regulatory negative bags to enhance the semantic matching capabilities of the SRM. These bags are processed through SASD II and refined using box mining strategy, resulting in the refined \(box_{srm}\). \textbf{iii). Segment Branch}: We use the final selected pseudo box to obtain corresponding mask proposals through IoU matching. Given that these mask proposals are less precise than ground truth, the branch employs \textbf{Affinity Refinement} at both pixel and high-level semantic space to construct soft pseudo masks alongside mask proposals for enhanced segmentation supervision.
   }

\label{fig:framework}
\end{center} 
\vspace{-15pt}
\end{figure*}

\textcolor{black}{SAPNet leverages single-point prompts with vision foundation models like SAM~\cite{DBLP:SAM} for initial proposal generation, subsequently refining these candidates through \textbf{proposal selection} and \textbf{selection refinement} modules. SAPNet++ builds on SAPNet by incorporating two novel components$-$\textbf{spatial-aware self-distillation} and \textbf{affinity propagation}$-$into the multi-stage strategy, which iteratively creates high-quality pseudo-labels and enables robust end-to-end training. As illustrated in Fig.~\ref{fig:framework}, these four key components are detailed below:}

\textcolor{black}{\textbf{(1) Proposal Selection (Sec.~\ref{sec:PSM}).} This branch aims to select semantically relevant proposals from those generated by SAM. It utilizes a multiple instance learning based~\cite{DBLP:MIL} strategy, augmented with point distance guidance, to mitigate the ``group issue'' where adjacent instances are erroneously merged, ultimately yielding initial pseudo-boxes.}

\textcolor{black}{\textbf{(2) Selection Refinement (Sec.~\ref{Sec:srm}).} 
Building on the initial selection, this stage enhances proposal quality through two parallel mechanisms. Proposal refinement is guided by robust positive and negative bags, while a dedicated box mining strategy explicitly tackles the ``local issue'' by dynamically merging proposals. Both processes contribute to generating higher-quality pseudo-boxes.}

\textcolor{black}{\textbf{(3) Spatial-Aware Self-Distillation (Sec.~\ref{sec: SASD}).}
As semantic relevance alone does not guarantee optimal proposal quality, we employ Spatial-Aware Self-Distillation during selection and refinement. This approach assesses the spatial completeness of each proposal, thereby enhancing robustness against granularity ambiguity (part \emph{vs.} whole).}

\textcolor{black}{\textbf{(4) Affinity Refinement (Sec.~\ref{sec:affinity}).} 
We introduce multi-level affinity refinement to resolve boundary ambiguity by propagating global and local affinities across pixel and semantic spaces. This process yields refined, soft pseudo-masks which, alongside masks from selection refinement branch, jointly provide enhanced supervision to the segmentation network, specifically improving its boundary delineation performance.}
\textcolor{black}{During training, all components are utilized, only the segmentation branch is retained for inference, ensuring efficiency.}

\vspace{-3pt}
\subsection{SAPNet}
The inherent lack of scale information in point annotations presents a key limitation. When used as prompts for vision foundation models like SAM, this often leads to the generation of multiple candidate masks for the same object, exhibiting ambiguity in granularity (\eg, representing parts versus the whole). Directly utilizing these ambiguous candidates for segmentation supervision can introduce significant noise, potentially degrading model performance. Our objective is therefore to develop a category-specific segmentor that selectively leverages only the semantically most representative proposals to ensure robust supervision.
\subsubsection{Proposal Selection Mechanism}\label{sec:PSM}
\setlength{\parindent}{2em}
\textbf{Proposal Selection.}
\textcolor{black}{Inspired by~\cite{DBLP:wsddn,DBLP:cpf}, our proposal selection mechanism adopts multiple instance learning (MIL)~\cite{DBLP:MIL}, which groups object-specific proposals into bags and assumes at least one true object exists per positive bag, and leverages label information to prioritize high-confidence proposals for segmentation, as illustrated in Fig.~\ref{fig: psm}.}
Given an image \(\mathbf{I} \in \mathbb{R}^{H \times W}\) with a set of \( N \) point annotations \(\mathcal{Y}_n = \{ (p_i, c_i) \}_{i=1}^{N} \), where \( p_i \) denotes the spatial coordinate and \( c_i \) is the corresponding class index, we aim to predict instance masks for each point.
For each annotated point \( p_i \), we first generate \( M \) category-agnostic mask proposals using SAM~\cite{DBLP:SAM}. \textcolor{black}{These \( M \) masks are then converted into a bounding-box proposal bag \( \mathcal{B}_i \in \mathbb{R}^{M \times 4} \), where the dimension 4 corresponds to the bounding-box coordinates \([x, y, w, h]\). This conversion to bounding boxes significantly reduces computational requirements compared to using masks, while while retaining a spatial contextual information around the object location. }
\textcolor{black}{As illustrated in Fig.~\ref{fig:framework}, image features \( \mathbf{F}_i \in \mathbb{R}^{M \times H \times W \times D} \) are extracted for the proposal bag \( \mathcal{B}_i \) by processing the \( M \) proposals using $7\times7$ RoIAlign~\cite{DBLP:MASK-RCNN} followed by two fully-connected layers.}

Similar to \cite{DBLP:wsddn,DBLP:oicr}, the extracted features \( \mathbf{F} \) serve as input to parallel classification and instance branches. These two branches employ distinct fully-connected layers, denoted \( f_{cls} \) and \( f_{ins} \), to generate \( \mathbf{W}_{{cls}}  \in \mathbb{R}^{M \times K} \) and \( \mathbf{W}_{ins}  \in \mathbb{R}^{M \times K} \), respectively. 
Softmax activation is then applied: along the class dimension (\( K \)) for classification scores \( \mathbf{S}_{{cls}} \in \mathbb{R}^{M \times K} \), and along the instance dimension (\( M \)) for instance scores \( \mathbf{S}_{{ins}} \in \mathbb{R}^{M \times K} \), \ie,
\begin{equation}\footnotesize
\vspace{-8pt}
\begin{aligned}
&  \mathbf{W}_{cls} = f_{cls}(\mathbf{F}),
\; [\mathbf{S}_{cls}]_{mk} = e^{[\mathbf{W}_{cls}]_{mk}}\big/\sum\nolimits_{k=1}^{K} e^{[\mathbf{W}_{cls}]_{mk}}, \\
& \mathbf{W}_{ins} = f_{ins}(\mathbf{F}), \;
[\mathbf{S}_{ins}]_{mk} = e^{[\mathbf{W}_{ins}]_{mk}}\big/\sum\nolimits_{m=1}^{M} e^{[\mathbf{W}_{ins}]_{mk}},
\label{Eq:initial score}
\end{aligned} 
\end{equation}
where $[\cdot]_{mk}$ is the value at the $m$-th row and $k$-th column.

\vspace{2pt}
\textbf{Point Distance Guidance.}\label{sec:PDG}
\textcolor{black}{As illustrated for SAM’s granularity ambiguity in Fig.~\ref{fig:issues} and MIL’s aggregation of adjacent high-response areas (noted in \cite{DBLP:wsddn,DBLP:midn}), this approach tends to erroneously merge adjacent objects of the same category due to point granularity ambiguities, resulting in inflated scores for merged instances.}
To address this, we integrate the instance-level annotated point information and employ a spatially aware selection strategy featuring a point-distance penalty mechanism.

To mitigate the challenge of overlapping objects and enhance model optimization, we implement a strategy that penalizes overlaps.
\textcolor{black}{For each $m$-th proposal within the set $\mathcal{B}_i$, we assign $t_{mj} = 1$ if it overlaps with any proposal in another identical class object's proposal bag $\mathcal{B}_j$, and also contains a ground truth annotation point of the object represented by $\mathcal{B}_j$. Otherwise, $t_{mj} = 0$.
The imposed penalty increases with the distance between the overlapping object and the object corresponding to the target proposal}. In Fig.~\ref{fig: point}, this penalty \(\mathbf{W}_{dis}\), calculated as the Euclidean distance between the annotated points of the overlapping proposals, is then transformed, by passing the reciprocal of \(\mathbf{W}_{dis}\) through a sigmoid function to compute the distance score, \(\mathbf{S}_{dis}\), for the proposal.
\vspace{-8pt}
\begin{equation}\small
\vspace{-5pt}
\begin{aligned}
&[\mathbf{W}_{dis}]_{im}=\sum\limits_{j=1,j \neq i}^{N}\left\|p_i-p_j\right\| \cdot t_{mj},\\
&[\mathbf{S}_{dis}]_{im} = \operatorname{Sigmoid}([1/ \mathbf{W}_{dis}]_{im})^{-d},
\label{Eq:MIL score}
\end{aligned} 
\end{equation}
where $[\cdot]_{im}$ is the value at the row $i$ and column $m$ in the matrix, and $d>0$ is the exponential factor.

\textbf{PSM Loss.} \label{sec: psmloss.}
The final score \( \mathbf{S} \) of each proposal is obtained by computing the Hadamard product of the classification score, the instance score, and the distance score, while the score \(\widehat{\mathbf{S}}\) for each proposal bag \( B_i \) is obtained by summing the scores of the proposals in \( B_i \). The PSM loss is constructed using the form of binary cross-entropy, and it is defined as follows:
\vspace{-5pt}
\begin{equation}\footnotesize
\begin{aligned}
& \mathbf{S}=\mathbf{S}_{cls} \odot \mathbf{S}_{ins} \odot \mathbf{S}_{dis} \in \mathbb{R}^{M \times K}, \;\;
\widehat{\mathbf{S}}= \sum\limits_{m=1}^{M} [\mathbf{S}]_m \in \mathbb{R}^{K}, \\
& \mathcal{L}_{psm} = \operatorname{CE}(\widehat{\mathbf{S}}, \mathbf{c}) =-\frac{\mathrm{1}}{N}\sum\limits_{n=1}^{N}\sum\limits_{k=1}^{K} \mathbf{c}_k \log(\widehat{\mathbf{S}}_k) + (1-\mathbf{c}_k)  \log(1-\widehat{\mathbf{S}}_k),
\label{Eq:bce loss}
\end{aligned} 
\vspace{-5pt}
\end{equation}
where $\mathbf{c} \in \{0, 1\}^{K}$ is the one-hot category label.

\textcolor{black}{With the guidance of PSM loss, the proposal selection skillfully identifies each proposal's category and and its corresponding instance.} The method selects the proposal with the highest score, marked as $ \mathbf{S} $, for a specific object as the pseudo box for the initial stage.

\subsubsection{Selection Refinement Mechanism}\label{Sec:srm}

In the initial proposal selection phase, we use enhanced MIL to select high-quality proposals from the positive bag \(\mathcal{B}^+\). However, as shown in Fig.~\ref{fig:framework}, the single-stage approach often yields suboptimal results prone to local optima. Inspired by PCL~\cite{DBLP:pcl}, we introduce a two-stage refinement strategy. Unlike typical WSOD methods relying solely on classification signals for refinement, our approach establishes high-quality positive and negative bags and integrates classification and instance branches to refine proposals, directly improving bounding box quality.

{\textbf{Positive and Negative Proposals Generator.}}
Following selection through proposal selection mechanism (PSM), we utilize the Positive and Negative Proposals Generator to enhance box proposal quality. Positive Proposals Generator aims to enrich the positive sample set \(\mathcal{B}^+\) and the overall bag quality. Negative Proposals Generator focuses on generating negative samples, including each object's background and part samples, which are essential for addressing local issues and refining box proposal selection.

\textbf{(a)} Positive Proposals Generator. Within this phase, coupled with the point distance penalty score \( \mathbf{S}_{{dis}} \) attributed to each proposal, we capitalize on the $\boldsymbol{b}_{psm}$ derived from the PSM stage to implement adaptive sampling for the identified bounding box. To further elaborate, for each $\boldsymbol{b}_{psm}$ (denoted as \( b_x^* \), \( b_y^* \), \( b_w^* \), \( b_h^* \)) isolated during the PSM phase, its dimensions are meticulously recalibrated leveraging a scale factor $v$ and its associated within-category inclusion score \( \mathbf{S}_{{dis}} \) to generate an augmented set of positive proposals \( (b_x, b_y, b_w, b_h) \). The equation is as follows:
\begin{equation}\small
\vspace{-3pt}
\begin{aligned}
b_w = (1 \pm v / \mathbf{S}_{dis}) \cdot b^*_w, \quad b_h = (1 \pm v / \mathbf{S}_{dis}) \cdot b^*_h, \\
b_x = b^*_x \pm (b_w - b^*_w)/2 , \quad b_y = b^*_y \pm (b_h - b^*_h)/2.
\end{aligned}
\label{Eq: PRB sampling}
\vspace{-3pt}
\end{equation}
The newly cultivated positive proposals are seamlessly integrated into the existing set \(\mathcal{B}_i\) to enrich the pool of positive instances. This enhancement is crucial for optimizing the training of the subsequent selection refinement.

\textbf{(b)} Negative Proposals Generator. MIL-based selection within a single positive bag may overemphasize the background noise, leading to inadequate focus on the whole object. To solve this, we create a negative bag from the background proposals post-positive bag training, which helps MIL maximize the attention toward the object.
Considering the image dimensions, we randomly sample proposals according to each image's width and height, for negative instance sampling. We assess the IoU between these negative and positive sets, filtering out those below $T_{neg1}$.

Additionally, based on its width and height, we enforce the smaller proposals' sampling with an IoU under a second threshold, $T_{neg2}$, from the inside $\boldsymbol{b}_{psm}$, which highest scored in PSM, as negative examples to rectify the MIL localization errors (local issues). These negative instances, partially capturing the object, drive the model to select high-quality bounding boxes, encompassing the entire object. 
The method is systematically elaborated in Algorithm~\ref{Alg: positive and negative proposals generation (pnpg)}.
\begin{algorithm}[tb!]
{
\small
\caption{Positive and Negative Proposals Generation}
\label{Alg: positive and negative proposals generation (pnpg)}
\textbf{Input:} $T_{neg1}$, $T_{neg2}$, $\boldsymbol{b}_{psm}$ from proposal selection stage, image $\mathbf{I}$, positive bags~$\mathcal{B}^+$. \\
\textbf{Output:} Positive proposal bags $\mathcal{B}^+$, Negative proposal set $\mathcal{U}$.\\
\vspace{-10pt}
\begin{algorithmic}[1]
\STATE \texttt{\footnotesize \# Step1: positive proposals sampling}
\FOR{$i = 1$ to $N$}
\STATE ${\mathcal{B}^{+}_i} \leftarrow \mathcal{B}_i$, $\mathcal{B}_i \in \mathcal{B}$;
\STATE ${\mathcal{B}^{+}_i} = {\mathcal{B}^{+}_i}\ \bigcup \operatorname{PosSample}(\boldsymbol{b}_{psm}^i)$;
\ENDFOR
\STATE \texttt{\footnotesize \# Step2: background negative proposals sampling}
\STATE $\mathcal{U} \leftarrow \emptyset$;
\STATE $\hat{\mathcal{B}} \leftarrow $ $\operatorname{RandSample}(\mathbf{I})$; 
\STATE $ {iou} = \operatorname{IoU}(\hat{\mathcal{B}}, \mathcal{B}_i)$ for each $\mathcal{B}_i \in \mathcal{B}$;
\IF{$ {iou}< T_{neg1}$ } 
 \STATE $\mathcal{U} = \mathcal{U}\ \bigcup \hat{\mathcal{B}}$;
\ENDIF
\STATE \texttt{\footnotesize \# Step3: part negative proposals sampling}
\FOR{$i = 1$ to $N$}
\STATE $\hat{\mathcal{B}} \leftarrow \operatorname{PartNegSample}(\boldsymbol{b}_{psm}^i)$ ;
\STATE ${iou} = \operatorname{IoU}(\hat{\mathcal{B}}, \boldsymbol{b}_{psm}^i)$ ;
\IF{ ${iou} < T_{neg2}$ } 
 \STATE $\mathcal{U} = \mathcal{U}\ \bigcup\ \hat{\mathcal{B}}$;
\ENDIF
\ENDFOR
\end{algorithmic}
}
\end{algorithm}

\begin{figure}[tb!]
\centering

\begin{subfigure}[t]{1.\linewidth}
    \centering
    \includegraphics[width=1.\linewidth]{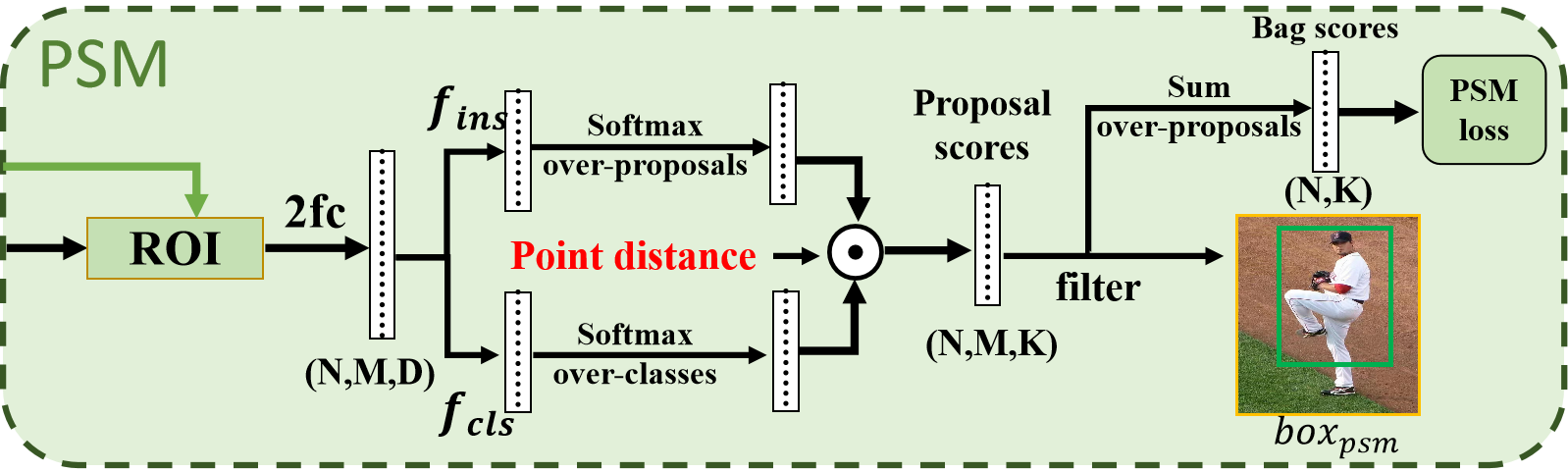}
    \vspace{-12pt}
    \caption{Proposal selection mechanism.}
    \label{fig: psm}
\end{subfigure}

\vspace{3pt} 
\begin{subfigure}[t]{1.\linewidth}
    \centering
    \includegraphics[width=1.\linewidth]{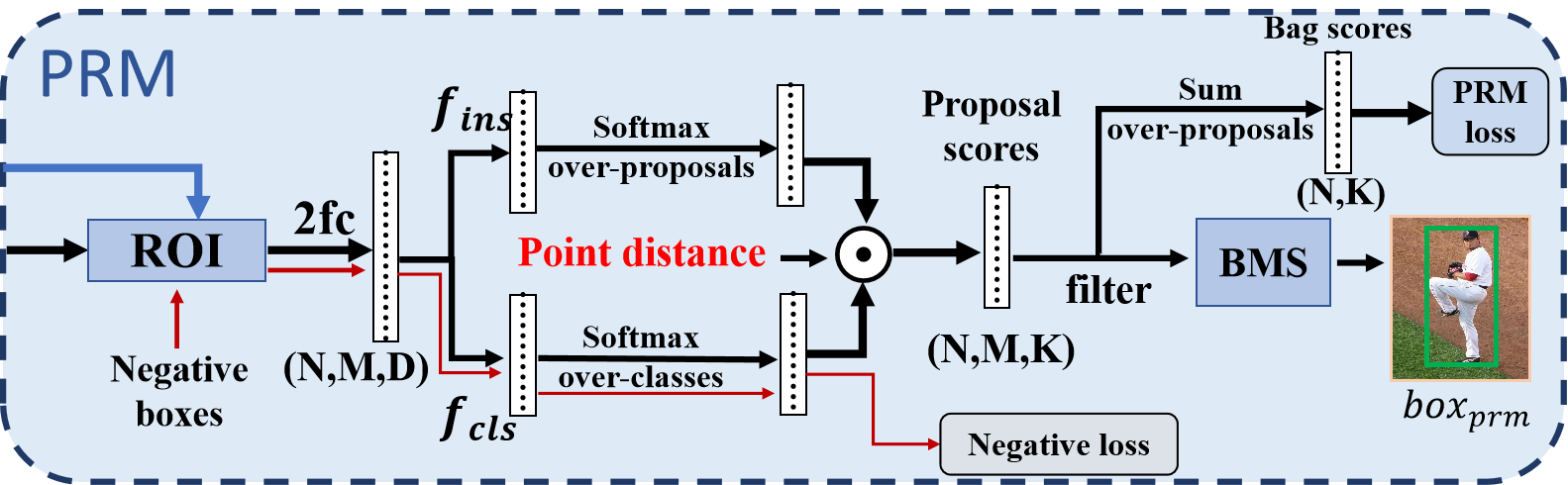}
    \vspace{-12pt}
    \caption{Selection refinement mechanism.}
    \label{fig: srm}
\end{subfigure}

\vspace{3pt} 
\begin{subfigure}[t]{1.\linewidth}
    \centering
    \includegraphics[width=1.\linewidth]{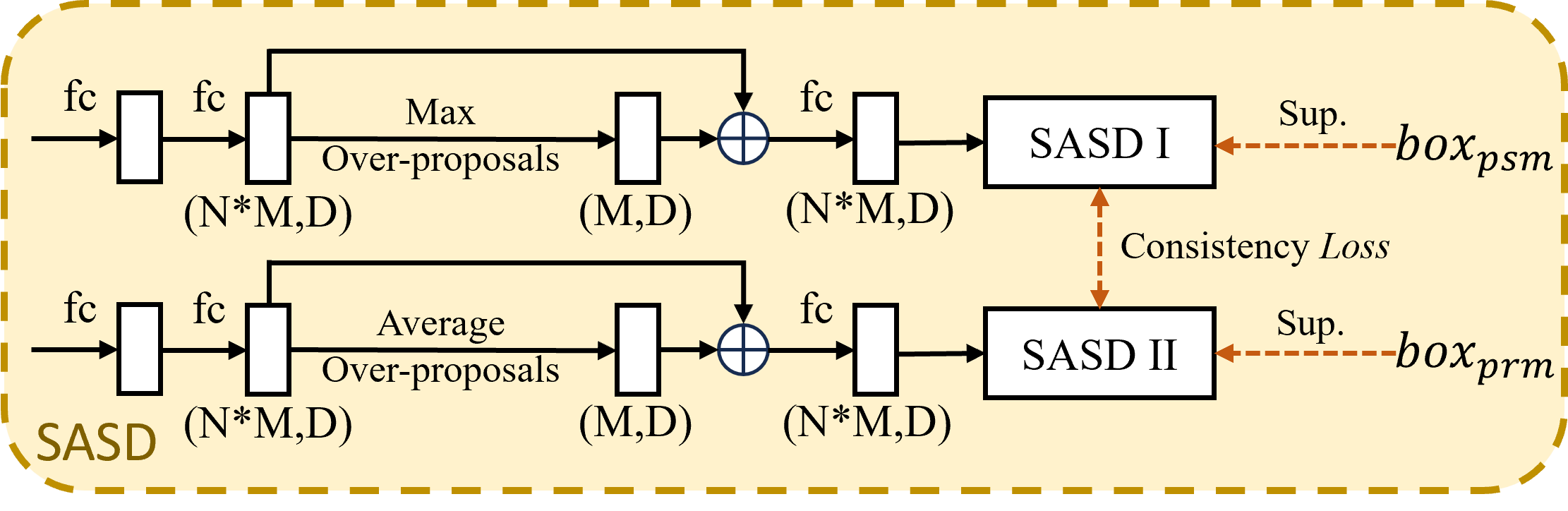}
    \vspace{-12pt}
    \caption{Spatial-aware self-distillation.}
    \label{fig: SASD}
\end{subfigure}

\caption{(a) The PSM weights scores from category, instance, and point-guided strategies to preliminarily filter proposals. (b) The SRM further enhances the selection capability by introducing negative examples and employs box mining strategy to address the local issues. (c) The SASD predicts a completeness score for each proposal in PSM and SRM, re-weighting them with the previous scores.}
\label{fig:combined}
\vspace{-10pt}
\end{figure}

\textbf{Selection Refinement.}
As is illustrated in Fig.~\ref{fig: srm}, the selection refinement extends the proposal selection's capabilities by focusing on selection and refinement. It merges positive instances from the positive proposal generator into the initial set to create an enriched \(\mathcal{B}^+\). It also integrates the negative instance set \(\mathcal{U}\) from negative proposal generator to provide a solid base for selection refinement. This integration modifies the MIL loss in selection refinement, substituting the CELoss with the Focal Loss for positive instances. The modified positive loss function is as follows:
\begin{equation}\small
\begin{aligned}
\setlength\abovedisplayskip{0pt}
\setlength\belowdisplayskip{1pt}
\mathcal{L}_{pos}  = \frac{1}{N}\sum\limits_{i=1}^{N} \left< \mathbf{c}^{\mathrm{T}}_i, \mathbf{\widehat{S}}_i \right> \cdot \operatorname{FL}(\mathbf{\widehat{S}^{*}}_i, \mathbf{c}_i),
\label{Eq:L_{bag}}
\end{aligned}
\end{equation}where $\operatorname{FL}$ is the focal loss~\cite{DBLP:retinanet_focalloss}, $\mathbf{\widehat{S}^*}_i$ and $\mathbf{\widehat{S}}_i$ represent the bag score predicted by SRM and PSM, respectively. ${\small{\left< \mathbf{c}^{\mathrm{T}}_i, \mathbf{\widehat{S}}_i \right>}}$ represents the inner product of the two vectors, indicating the predicted bag score of the ground-truth category. 

To enhance background suppression, we use negative proposals and introduce a dedicated loss for these instances. Notably, these negative instances pass only through the classification branch for instance score computation, with their scores derived exclusively from classification. The specific formulation of this loss function is detailed below:
\begin{equation}\small
\begin{aligned}
\setlength\abovedisplayskip{1pt}
\setlength\belowdisplayskip{1pt}
\beta =\frac{1}{N}\sum\limits_{i=1}^{N} \left< \mathbf{c}^{\mathrm{T}}_i, \mathbf{\widehat{S}}_i \right>,
\label{Eq:neg_pbr1}
\end{aligned}
\end{equation}
\vspace{-3pt}
\begin{equation}\small
\begin{aligned}
\setlength\abovedisplayskip{1pt}
\setlength\belowdisplayskip{1pt}
\mathcal{L}_{neg} = - \frac{1}{\left|\mathcal{U}\right|}\sum\limits_{\mathcal{U}}\sum\limits_{k=1}^{K} \beta \cdot ([\mathbf{S}^{cls}_{neg}]_k)^{2} \cdot \log(1-[\mathbf{S}^{cls}_{neg}]_k).
\label{Eq:neg_pbr2}
\end{aligned}
\end{equation}

The SRM loss consists of the $\mathcal{L}_{pos}$ for positive bags and the negative loss $\mathcal{L}_{neg}$ for negative samples, \ie,
\begin{equation}
\begin{aligned}
\mathcal{L}_{srm} = \mathrm{\alpha} \cdot \mathcal{L}_{pos}  + (1-\mathrm{\alpha}) \cdot \mathcal{L}_{neg},
\label{Eq:PBR basic loss}
\end{aligned}
\end{equation}
\noindent where $\mathrm{\alpha}=0.25$ by default.

\textbf{Box Mining Strategy.}
Under single point prompts, MIL's preference for segments with significant foreground presence and SAM's tendency to capture only object parts often result in the final bounding boxes, \(\boldsymbol{b}_{srm}\), which inadequately cover the instances. To enhance bounding box quality, we introduce a box mining strategy in SRM that adaptively expands \(\boldsymbol{b}_{sel}\) from selection refinement by merging it with the original proposal filter. This approach aims to address MIL's localization challenges effectively.

\textcolor{black}{The Box Mining Strategy comprises two main components:
\textbf{i)} We select the top-\( k \) boxes according to SRM's scores from the positive proposal bag \( \mathcal{B}^+ \) to form a set ${\hat{\mathcal{B}}^+}$. Boxes in ${\hat{\mathcal{B}}^+}$ (denoted as \(\hat{\boldsymbol{b}}^+\)) are evaluated against \( \boldsymbol{b}_{sel} \) based on IoU and size criteria, with a minimum threshold \( T_{min1} \).}

\textcolor{black}{$\hat{\boldsymbol{b}}^+$ exceeding \( \boldsymbol{b}_{sel} \) in size and meeting the IoU threshold undergo a dynamic expansion. This process, which considers IoU adjustments, allows for adaptive integration with \( \boldsymbol{b}_{sel} \), effectively addressing the local issue and improving boundary localization.
\textbf{ii)} Challenges associated with locality often result in an exceedingly low IoU between \(\hat{\boldsymbol{b}}^+\) and \( \boldsymbol{b}_{sel} \). However, the ground truth may fully encompass \( \boldsymbol{b}_{part} \). In cases where the conditions from component (i) are not met, if a \(\hat{\boldsymbol{b}}^+\) can completely encapsulate \( \boldsymbol{b}_{sel} \), we reset the threshold \( T_{min2} \). \(\hat{\boldsymbol{b}}^+\) that exceed this threshold are adaptively merged with \( \boldsymbol{b}_{sel} \) to form the final \( \boldsymbol{b}_{srm} \), which is then used to generate \( \mathcal{M}_{srm} \). These components together establish our box mining strategy. The specifics of this approach are outlined in Algorithm~\ref{Alg:BMS}.}
\begin{algorithm}[tb!]
{
\small
\caption{Box Mining Strategy}
\label{Alg:BMS}
\textbf{Input:} $T_{min1}$, $T_{min2}$, $\boldsymbol{b}_{{sel}}$ from selection refinement, initial positive bag $\mathcal{B}^+$, image $\mathbf{I}$, $N$ is the number of objects in $\mathbf{I}$. \\
\textbf{Output:} Final bounding box for selection refinement $\boldsymbol{b}_{srm}$.\\
\vspace{-10pt}
\begin{algorithmic}[1]
\FOR{$i \in N$}
\STATE ${\hat{\mathcal{B}}_i} \leftarrow \operatorname{TopK}(\mathcal{B}_i, k)$, where $\mathcal{B}_i \in \mathcal{B}^+$;
\STATE ${cnt} = 0$;
\IF {$\boldsymbol{b}_j^w \cdot \boldsymbol{b}_j^h > \boldsymbol{b}_{sel}^h \cdot  \boldsymbol{b}_{sel}^w$ for each $\boldsymbol{b}_j \in {\hat{\mathcal{B}}_i}$}
\STATE $iou = \operatorname{IoU}(p_j, \boldsymbol{b}_{sel})$;
\IF{ $iou > T_{min1}$ } 
\STATE $\boldsymbol{b}_{srm} \leftarrow \left( \boldsymbol{b}_j + iou \cdot \boldsymbol{b}_{{sel}} \right) / \left( iou + 1 \right)$;
\STATE $T_{min1} = iou$
\STATE $cnt=cnt+1$;
\ELSIF {{$cnt=0$ and $\boldsymbol{b}_{sel} \in \boldsymbol{b}_j$ and $iou > T_{min2}$}}
\STATE $\boldsymbol{b}_{srm} \leftarrow (\boldsymbol{b}_j \cdot iou + \boldsymbol{b}_{sel}) / (iou+1)$;
\STATE $T_{min2} = iou$
\ENDIF
\ENDIF
\ENDFOR
\end{algorithmic}
}
\end{algorithm}

\subsection{SAPNet++}\label{sec:sasd}

Based on SAPNet, the proposal selection branch only relies on semantic distinguish ability to choose the proposal that has the best semantic representation. However, the proposal with the best classification score might not have the most complete mask, conflicts with the requirements of instance segmentation. Therefore, we propose spatial-aware self-distillation to enhance the proposal selection. Further, due to the limitation of SAM's mask quality in specific tasks, the best selected mask may still have a significant gap from the actual ground-truth. The multi-level affinity propagation is utilized to refine the estimated pseudo mask. These designs, along with the basis SAPNet, yield the newly proposed SAPNet++. 

\subsubsection{Spatial-aware Self-distillation for Proposal Selection}\label{sec: SASD}

The fundamental proposal selection mechanism relies on classification confidence, which is theoretically flawed: due to \textbf{granularity ambiguity}, a semantically discriminative part (e.g., "a shirt") may receive a higher classification score than a complete instance (e.g., "a person"). This leads the model to converge to a suboptimal solution by selecting spatially incomplete pseudo-labels. To fundamentally address this issue, we propose \textbf{Spatial-aware Self-distillation (SASD)}, a self-supervised learning framework designed to explicitly model and optimize for \textbf{spatial completeness} in the absence of direct supervision.

Our core idea is that an ideal pseudo-label should be not only semantically correct but also spatially complete. We formally define the \textbf{completeness} of a proposal (or mask) \(m\) with respect to its true, yet unknown, full instance mask \(M^{gt}\) as a function \(\Omega(m, M^{gt}) \in [0, 1]\), whose value is maximized when \(m\) perfectly covers \(M^{gt}\). However, in our weakly-supervised setting, \(M^{gt}\) is an unobservable \textbf{latent variable}.
SASD provides a principled solution to this latent variable estimation problem. We design a completeness predictor, denoted as a function \(\phi_{comp}\) with parameters \(\boldsymbol{\xi}\), that aims to predict a completeness score for any given proposal \(m\):
\begin{equation}
    s_{comp}(m) = \phi_{comp}(m; \boldsymbol{\xi}).
\end{equation}
The supervision signal for learning \(\phi_{comp}\) originates not from external annotations but from the model's own best estimate at the current training stage, creating a self-distillation feedback loop.

\textbf{Context-aware Feature Enhancement.} To enable the completeness predictor to make more accurate judgments, we first enhance the context-awareness of its input features. As illustrated in Fig.~\ref{fig: SASD}, for a proposal bag \(\mathcal{B}_i\), we obtain its initial features \(\mathbf{F}_i\) via RoI alignment and two fully connected (fc) layers. We then average the features of the top-\(k\) proposals within the bag to obtain a feature \(\mathbf{F}_i^+ \in \mathbb{R}^{1 \times D}\) that represents the core context of the object. This context feature is then broadcast to \(\mathbb{R}^{M \times D}\) and added to each proposal's feature to yield the enhanced features \(\mathbf{F}_i^*=\mathbf{F}_i+\mathbf{F}_i^+\), which fuse both local and global information.

\textbf{Self-distillation for Spatial Completeness.} The completeness predictor \(\phi_{comp}\) is implemented as an fc layer that maps the enhanced features \(\mathbf{F}_i^*\) to a predicted completeness score vector $[\mathbf{S}_{comp}]_i \in \mathbb{R}^{M \times 1}$. The key to SASD is using the best pseudo-box \(\hat{b}^*_i\) selected by the model at the current stage as a proxy for the unknown latent variable \(M^{gt}\). The learning objective for the predictor's parameters \(\boldsymbol{\xi}\) is to minimize the discrepancy between its predictions and this self-generated target, which can be formally expressed as:
\begin{equation}
    \boldsymbol{\xi}^* = \arg\min_{\boldsymbol{\xi}} \sum_{i=1}^N \sum_{m_j \in \mathcal{B}_i} \mathcal{L}_{distill}(\phi_{comp}(m_j; \boldsymbol{\xi}), \Omega(m_j, \hat{b}^*_i)),
\end{equation}
where \(\mathcal{L}_{distill}\) is a distillation loss. Subsequently, we \textbf{instantiate} the abstract completeness function \(\Omega\) as a concrete, computable metric: the \textbf{Intersection over Union (IoU)}. This provides a self-generated supervision target \(T_i = \text{IoU}(\mathcal{B}_i, \hat{b}^*_i)\).

In practice, we implement this distillation process using the following loss function, which serves as the concrete form of \(\mathcal{L}_{distill}\). For optimization stability, the self-generated supervision target \(T_i\) is linearly normalized to \(T^{\prime}_i=2T_i-1 \in (-1,1)\).
\begin{equation}\small
    \begin{aligned}
    \mathcal{L}_{sasd} ={} & \operatorname{SL_1}([\mathbf{S}_{comp}]_i,T^{\prime}_i)
    + \operatorname{SL_1}([{\mathbf{S}}^*_{comp}]_i,T^{*\prime}_i) \\
    +{} & \operatorname{SL_1}([\mathbf{S}_{comp}]_i,[{\mathbf{S}}^*_{comp}]_i),
    \end{aligned}
    \label{eq:Comp_loss}
\end{equation}
where \(\operatorname{SL_1}\) denotes the Smooth \(L_1\) loss. This loss function drives the model to learn from its own "knowledge" (the first term) and stabilizes the learning process via consistency regularization (the latter two terms). 

Finally, the normalized completeness score $[\mathbf{S}_{comp}]^{\prime}_i$ is multiplied by the semantic confidence score \(\mathbf{S}_i\) to produce a \textbf{holistic suitability score} \(\mathbf{S}^{\prime}_i\) that balances both semantic and spatial qualities:
\begin{equation}
    \mathbf{S}^{\prime}_i = [\mathbf{S}_{comp}]^{\prime}_i \odot \mathbf{S}_i.
\end{equation}

This holistic score is then used to guide the selection of a more spatially complete pseudo-box \(\hat{b}^*\) in the next iteration:
\begin{equation}
    \hat{b}^* = \arg\max_{m \in \mathcal{B}_i} [\mathbf{S}^{\prime}_i]_m.
\end{equation}

This creates a virtuous cycle of \textbf{"better completeness predictor \(\rightarrow\) more complete pseudo-label \(\rightarrow\) higher-quality self-supervision signal",} which progressively improves the estimation of the latent variable.

\subsubsection{Multi-level Affinity Refinement}\label{sec:affinity}
The segmentation results supervised by pseudo labels from selection refinement mechanism (SRM) are generally coarse, partly because SRM does not guarantee optimal outcomes due to inherent granularity ambiguity in point annotations. Additionally, the quality of proposals generated by current proposal generators via single-point prompts often deviates significantly from the ground truth, presenting \textbf{``boundary uncertainty''} such as holes and irregular boundaries. Therefore, our primary focus is on further refining these segmentation results. We provide a further detailed analysis of these specific limitations in Appendix B.

\begin{figure}[tb!]
\begin{center}
     \includegraphics[width=1.\linewidth]{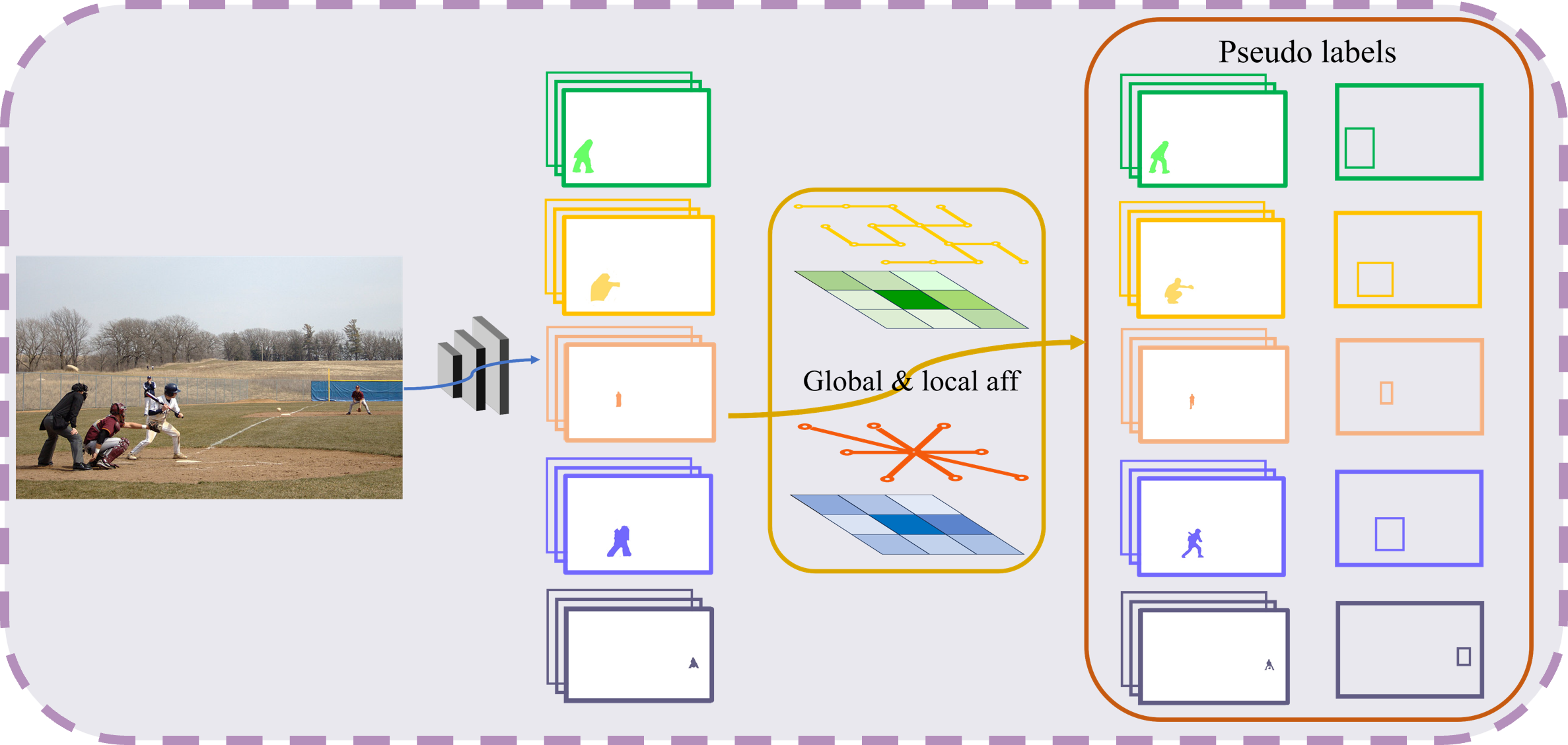}
\vspace{-10pt}
\caption{Multi-level Affinity Refinement integrates pixel-level color and texture information with high-dimensional semantic information from the backbone, refining segmentation masks through global and local affinity.}
\label{fig: aff}
\end{center}
\vspace{-15pt}
\end{figure}

Inspired by recent weakly-supervised segmentation approaches\cite{DBLP:Boxinst,DBLP:afa,DBLP:apro}, which predominantly consider generating pseudo labels and then leveraging weak annotations, such as boxes, points, or image-level labels, to supervise the segmentation network for producing high-quality segmentation outcomes, we propose to introduce multi-level affinity on top of network predictions to generate high-quality pseudo masks, as is illustrated in Fig.~\ref{fig: aff}.

Specifically, we construct unary and pairwise potentials at the low-level image and high-level semantic feature layers. In the conditional random field framework, unary potentials represent the pixel‘s confidence assigned labels, while pairwise potentials can reflect pairwise relationships such as distance and color similarity among pixels at the low-level image. We model the affinity relationships between pixels using pairwise potentials and propagate this through the unary potentials.
Additionally, we extend this method to multi-scale semantic features to capture more advanced semantic affinity. Utilizing low-level pixel and high-level semantic affinity, we generate high-quality pseudo labels, which can be defined as follows:
\begin{equation} \small
\vspace{-5pt}
\begin{aligned}
{{Y}^I_i} & = \frac{1}{{{z_i}}}\sum\limits_{{j} \in \tau} {\phi ({x_j})} \psi ({x_i},{x_j}),\\
{{Y}^S_i} & = \frac{1}{{{u_i}}}\sum\limits_{{j} \in \tau} {\phi ({x_j})} {\psi}^{\prime} ({x_i},{x_j}),
\end{aligned}
\label{equ:low-level}
\end{equation}
where, ${\phi}({x_j})$ represents the unary potential term, aligning $y_i$ and \(\mathcal{M}_{pre}\) from network prediction based on point prompts. $\psi ({x_i},{x_j})$ and ${\psi}^{\prime} ({x_i},{x_j})$ denote the pairwise potential term, incorporating inter-pixel relationships and high-level semantic information to guide predictions and produce accurate pseudo labels. $\tau$ defines the region with varying receptive fields (kernel size). $z_i$ and $u_i$ denote the summation of pairwise affinity $\psi ({x_i},{x_j})$ and ${\psi}^{\prime} ({x_i},{x_j})$ respectively, normalized with respect to $j$.

We utilize the aforementioned method Equ.~\ref{equ:low-level} to generate high-quality pseudo labels by incorporating global affinity and local pairwise affinity on the original image and semantics feat. 
Next, we assign each $Y_i$ with \(\mathcal{M}_{pre}\) and directly employ the $L_1$ distance as the objective for unlabeled regions.

\textbf{Global Affinity Refinement.}
We initially model global affinity relations on the input image and multi-scale semantic features to capture long-range affinity relationships while maintaining spatial consistency across the image. We establish a 4-connected planar graph $\mathcal{G}$, where each node is adjacent to up to four neighbors. This structure is preferred over an 8-connected graph to reduce the computational overhead and minimize the edge noise in long-distance affinity computation. The weight of each edge represents the pixel distance (or the semantic distance in high-level features) between adjacent nodes, encapsulating the geometric and semantic proximity essential for robust image segmentation.

Motivated by \cite{DBLP:treeloss,DBLP:learnabletree}, which utilizes the minimal spanning tree (MST) properties to preserve long-range dependencies and fine details while reducing computational overhead, we adopt a tree filter approach. This method filters out the longest edges on graph $\mathcal{G}$, resulting in a sparser graph structure for the input image ($\mathcal{G}_I$) and semantic features ($\mathcal{G}_S$):
For the low-level image, $\mathcal{G}_I = {\mathcal{V}_I, \mathcal{E}_I}$, where $\mathcal{V}_I = {\mathcal{V}_i}^{{N}_I}$ is the set of nodes and $\mathcal{E}_I = {\mathcal{E}_i}^{{N}_I - 1}$ is the set of edges.
For high-level semantic features, $\mathcal{G}_S = {\mathcal{V}_S, \mathcal{E}_S}$, where $\mathcal{V}_S = {\mathcal{V}_i}^{{N}_S}$ is the set of nodes and $\mathcal{E}_S = {\mathcal{E}_i}^{{N}_S - 1}$.

We model global pairwise potentials by iterating through each node, designating the current node as the root of the generating graph $\mathcal{G}_I$, and propagating its long-term affinity relationships to other nodes. To establish affinity relations with distant nodes, we traverse along the minimal spanning tree path via neighboring nodes. Inspired by APro \cite{DBLP:apro}, we adopt the distance-insensitive maximum affinity function, which decays slowly during long-distance affinity propagation and minimizes over-segmentation by maximizing the difference between two nodes along the path. Consequently, we define the global pairwise potential ${\psi _I}^g$ and ${\psi _S}^g$, as follows:
\begin{equation}\small
  I = \operatorname{Bilinear}\left(\mathbf{I}, \mathcal{M}_{pre}.size()\right)
  \label{eq:imagesampling}
\end{equation}
\begin{equation}\small
  F = \operatorname{Bilinear}\left(\mathbf{F}, \mathcal{M}_{pre}.size()\right)
  \label{eq:featuresampling}
\end{equation}
\begin{equation}
{{\psi _I}^g}({x_i},{x_j}) = \mathop {\mathcal{G_I}({I_i},{I_j})}\limits_{\forall j \in \mathcal{V_I}}  = \exp ( - \mathop {\max }\limits_{\forall(k,t) \in {\mathbb{E}_{i,j}}} \frac{{{w_{k,t}}}}{{{\zeta_g}^2}}),
\end{equation}
\begin{equation}
{{\psi _S}^g}({x_i},{x_j}) = \mathop {\mathcal{G_S}({F_i},{F_j})}\limits_{\forall j \in \mathcal{V_S}}  = \exp ( - \mathop {\max }\limits_{\forall(k,t) \in {\mathbb{T}_{i,j}}} \frac{{{w'_{k,t}}}}{{{\sigma_g}^2}}),
\end{equation}
where $\mathcal{G_I}$ and $\mathcal{G_S}$ represent the global trees constructed on the original image and semantic feature space, respectively. $\mathbb{E}_{i,j}$ and $\mathbb{T}_{i,j}$ refer to the set of edges along the path of $\mathcal{G}$ from node $j$ to node $i$. The edge weight $w_{k,t}$ between the adjacent nodes $k$ and $t$  is viewed as the Euclidean distance between the pixel values of two adjacent nodes, \ie, ${w_{k,t}} = {\left| {{I_k} - {I_t}} \right|^2}$. The parameters $\zeta_g$ and ${\sigma_g}$ are employed to modulate the affinity with respect to long-range pixels, thereby controlling the influence of spatially distant nodes on the computed potentials.
In this way, utilizing global affinity on origin image and semantic feat to obtain the pseudo mask $I^g$ and $F^g$ can be formulated as follows: 
\begin{equation}\small
\begin{aligned}
{I_i^g} = \frac{1}{{{z_i^g}}}\sum\limits_{ j \in \mathcal{{V}_I}} {\phi ({x_j})} {{\psi_I}^g}({x_i},{x_j}), \  {z_i^g} = \sum\limits_{j \in {\mathcal{{V}_I}}} {{\psi _I}^g({x_i},{x_j})}_{[{x_i} \neq {x_j}]}, 
\end{aligned}
\end{equation}
\begin{equation}\small
\begin{aligned}
{F_i^g} = \frac{1}{{{u_i^g}}}\sum\limits_{ j \in \mathcal{{V}_S}} {\phi ({x_j})} {{\psi_S}^g}({x_i},{x_j}), \ {u_i^g} = \sum\limits_{j \in {\mathcal{{V}_S}}} {{\psi _S}^g({x_i},{x_j})}_{[{x_i} \neq {x_j}]},
\end{aligned}
\end{equation}
where interactions with distant nodes are facilitated through a tree-based topology, leveraging the structural advantages of the tree-based graph while maintaining topological consistency. 

Additionally, propagating affinity in low-level image and high-level semantic space ensures spatial consistency in color, texture, and semantics. This dual-level propagation strategy effectively aligns and harmonizes the perceptual attributes throughout the image, enhancing the overall coherence and quality of the pseudo masks.

\textbf{Cascade Local Affinity Refinement.}
Global affinity modeling on entire images and semantic feature maps enables the capture of long-range pixel and semantic relationships, which is crucial for maintaining consistency across the image and semantic spaces. However, due to the extensive propagation distance, pairwise affinity decreases, inevitably introducing noise such as adhesion between masks of adjacent instances of the same class and reducing the ability to handle fine details.

To address the edge noise introduced by global affinity and enhance the ability to process local details, thus improving mask smoothness, we introduce local affinity refinement. We use the higher-quality pseudo labels generated by global affinity as the input for local affinity refinement to further enhance results. Unlike global affinity, we employ a Gaussian kernel to establish dense affinity connections, enhancing local processing capabilities. This approach mitigates the global affinity propagation's shortcomings and strengthens the model's ability to discern and refine intricate details within the image.
The local pairwise term $\psi _s$ is defined as: 
\begin{equation}\small
{{\psi _I}^l}({x_i},{x_j}) = \mathop {{{\cal K}_I}{\rm{ }}({{I^g}_i},{{I^g}_j})}\limits_{j \in {{\cal N}_I}(i)}  = \exp \left( {\frac{{ - {{\left| {{{I^g}_i} - {{I^g}_j}} \right|}^2}}}{{\zeta_l ^2 }}} \right),
\end{equation}
\begin{equation}\small
{{\psi _S}^l}({x_i},{x_j}) = \mathop {{{\cal K}_S}{\rm{ }}({{F^g}_i},{{F^g}_j})}\limits_{j \in {{\cal N}_S}(i)}  = \exp \left( {\frac{{ - {{\left| {{{F^g}_i} - {{F^g}_j}} \right|}^2}}}{{\sigma_l ^2 }}} \right),
\end{equation}
where ${\cal K}_I$ and ${\cal K}_S$ denote the Gaussian kernel, $\mathcal{N}_I(i)$ and $\mathcal{N}_S(i)$ are the set containing all local neighbor pixels. 

\emph{\textbf{(a) Multi-scale Kernel.}} For multi-scale semantic features, due to the significant disparities in resolution across different scales, we tailor each scale's kernel size for to optimize affinity calculations. Specifically, we employ larger kernel sizes on high-resolution features to capture affinity over longer distances, enhancing the model's ability to integrate contextual information from broader areas. Conversely, on low-resolution features, we utilize smaller kernel sizes to minimize the introduction of noise while still acquiring useful affinity information. 

By adjusting the kernel size according to the resolution of the semantic features, we balance between capturing extensive contextual relationships and maintaining the clarity and relevance of the affinity signals, thus effectively improving the segmentation accuracy and robustness.

The degree of similarity is controlled by the parameters $\zeta_l$ and $\sigma_l$. Then the pseudo mask $I^l$ and $F^l$ can be obtained via the following affinity computation: 
\begin{equation}\small
\begin{aligned}
{I^l_i} = \frac{1}{{{{z}_i^l}}}\sum\limits_{j \in {\mathcal{N}_I}(i)} {{\phi}({x_j})} {{\psi_I}^l}({x_i},{x_j}), \
{z_i^l} = \sum\limits_{j \in {\mathcal{N_I}}(i)} {{{\psi _I}^l}({x_i},{x_j})}_{[{x_i} \neq {x_j}]},
\end{aligned}
\end{equation}
\begin{equation}\small
\begin{aligned}
{F^l_i} = \frac{1}{{{{u}_i^l}}}\sum\limits_{j \in {\mathcal{N}_S}(i)} {{\phi}({x_j})} {\psi_S}^l({x_i},{x_j}), \ 
{u_i^l} = \sum\limits_{j \in {\mathcal{N}_S}(i)} {{\psi _S}^l({x_i},{x_j})}_{[{x_i} \neq {x_j}]},
\end{aligned}
\end{equation}

\emph{\textbf{(b) Cascade Block.}} 
In our approaches, each local affinity propagation process is considered a block. The output pseudo mask from each block is combined with the block's input through a weighted operation to serve as the input for the subsequent block.  
\begin{equation}
\begin{aligned}
{[I^l_{in}]_{n+1}} & = {\lambda}_I \cdot {[I^l]_{n}}+(1 - {\lambda}_I) \cdot{[I^l_{in}]_{n}}, \\
{[F^l_{in}]_{n+1}} & = {\lambda}_S \cdot{[F^l]_{n}}+(1 - {\lambda}_S) \cdot{[F^l_{in}]_{n}}
\end{aligned}
\end{equation}
Where, ${\lambda}_I$ and ${\lambda}_S$ are set to 0.5 empirically. By adopting this cascaded structure, we enhance detail resolution without introducing excessive noise, ensuring that the final pseudo mask is high quality. This strategy enables a progressive refinement of segmentation details while maintaining the integrity and quality of the pseudo labels.

\textbf{Affinity Loss.}
After the cascade of affinity iterations, the pseudo masks derived from the original image and high-dimensional semantic features, denoted as ${I^l}_i$ and ${F^l}_i$, can be further expressed as ${Y^I}_i$ and ${Y^S}_i$. Empirically, we adopt the Euclidean distance to measure the discrepancy between the pseudo masks and \(\mathcal{M}_{pre}\) to compute the affinity loss, which is defined as follows:
\begin{equation}\small
\begin{aligned}
\mathcal{L}_{aff}=\frac{1}{\mathcal{M}_{reg}}\sum\limits_{i \in {N}}{\{{|{Y^I}_i-{Y}_i|}+{|{Y^S}_i-{Y}_i|}\}}
\end{aligned}
\end{equation}
where, \({\mathcal{M}_{reg}}\) represents the sum of the foreground regions of the mask used to supervise the segmentation branch, and \({Y}_i \) denotes the label of \(\mathcal{M}_{pre}\).

\subsection{Training and Inference}\label{sec:loss}

\textbf{Training Loss.} We utilize SOLOv2\cite{DBLP:solov2} as our segmentation branch for its efficiency and use $\mathcal{M}_{srm}$ from the $\boldsymbol{b}_{srm}$ mapping for segmentation supervision. The loss function for the segmentation head has two components: a class classification loss \(L_{cls}\), for which we empirically adopt focal loss~\cite{DBLP:retinanet_focalloss}, and the instance segmentation loss that is formulated as follows:
\begin{equation}\small
    \begin{aligned}
      \mathcal{L}_{mask}= 
       \mathcal{L}_{Dice}(\mathcal{M}_{pre},\mathcal{M}_{srm}) \\
    \end{aligned}
\end{equation}

The aggregate loss function, $L_{total}$ can be articulated as:
\begin{equation}\small
    \begin{aligned}
      \mathcal{L}_{total}=
      \mathcal{L}_{mask}
      + \mathcal{L}_{cls}
      + \lambda \cdot \mathcal{L}_{psm} 
      + \mathcal{L}_{srm} 
      + \mathcal{L}_{aff}
    \end{aligned}
\end{equation}
\noindent where $\mathcal{L}_{mask}$ is the Dice Loss~\cite{DBLP:Diceloss}, $\mathcal{L}_{cls}$ is the Focal Loss\cite{DBLP:retinanet_focalloss}, and $\lambda$ is set as 0.25.

\textcolor{black}{\textbf{Inference.} In the inference of SAPNet++ (FPS=43, FLOPS=183G), only the segmentation branch remains active post-training. Given an input image, mask predictions are directly generated through an optimized Matrix Non-Maximum Suppression (Matrix-NMS) mechanism. Crucially, the additional modules introduced during training—including proposal selection, selection refinement, spatial-aware self-distillation, and affinity head—incorporate zero computational overhead during inference. These modules exclusively contribute to training efficiency improvements while maintaining complete cost neutrality in deployment scenarios, ensuring the model retains its native lightweight characteristics during real-time applications, more cost details in Tab. \textcolor{red}{19}}.

\section{Experiment}\label{sec:experiment}
\begin{table*}
\renewcommand\arraystretch{1.1}
\begin{center}
\resizebox{1.0\textwidth}{!}{
\begin{tabular}{l|c|c|c|c|ccc|ccc}
\specialrule{0.15em}{0pt}{2pt}  
Method & Ann.& Backbone & sched. &Arch. &$\rm mAP$ & $\rm mAP_{50}$ & $\rm mAP_{75}$ &$\rm mAP_{s}$ &$\rm mAP_{m}$ &$\rm mAP_{l}$\\
\specialrule{0.08em}{2pt}{2pt}

\rowcolor[rgb]{ 0.95,  0.95,  0.95} \multicolumn{11}{c}{\textit{\textbf{Fully-supervised instance segmentation models.}}\vspace{2pt}} \\

Mask R-CNN~\cite{DBLP:MASK-RCNN}&$\mathcal{M}$ &ResNet-50&1x&Mask R-CNN& 34.6 &56.5  &36.6 &18.3& 37.4& 47.2\\
YOLACT-700~\cite{DBLP:yolact}&$\mathcal{M}$ &ResNet-101&4.5x&YOLACT& 31.2 &54.0  &32.8 &12.1& 33.3& 47.0\\
PolarMask~\cite{DBLP:polarmask}&$\mathcal{M}$ &ResNet-101&2x&PolarMask& 32.1 &53.7  &33.1 &14.7& 33.8& 45.3\\
SOLO~\cite{DBLP:SOLO} &$\mathcal{M}$&ResNet-50&1x&SOLO&33.1& 53.5& 35.0&12.2& 36.1& 50.8\\
SOLOv2~\cite{DBLP:solov2} &$\mathcal{M}$&ResNet-50&1x&SOLOv2&34.8& 54.9& 36.9&13.4& 37.8& 53.7\\
CondInst~\cite{DBLP:condinst} &$\mathcal{M}$&ResNet-50&1x&CondInst&35.3& 56.4& 37.4&18.0& 39.4& 50.4\\
SwinMR~\cite{DBLP:swin-transformer} & $\mathcal{M}$ &Swin-S&50e&SwinMR&43.2& 67.0 &46.1 &24.8& 46.3& 62.1\\
IFormer~\cite{iformer} &$\mathcal{M}$&ViT-M&1x&Mask-RCNN & 37.9&59.7&40.7 &-& -& -\\
Mask2Former~\cite{DBLP:mask2former} &$\mathcal{M}$&ResNet-50&1x&Mask2Former & 38.7&59.8&41.2 &18.2& 41.5& 59.8\\
SAM-CLIP~\cite{sam-clip} &$\mathcal{M}$&ViT-B&39e&SAM & 40.9&-&- &-& -& -\\
MaskDINO~\cite{DBLP:maskdino} &$\mathcal{M}$&ResNet-50&1x&MaskDINO & 41.4&62.9&44.6 &21.1& 44.2& 61.4\\
Mask2Former~\cite{DBLP:mask2former} &$\mathcal{M}$&Swin-S&50e&Mask2Former & 46.1&69.4&52.8 &25.4& 49.7& 68.5\\
Semantic-SAM~\cite{semantic-sam} &$\mathcal{M}$&Swin-T&3x&MaskDINO & 47.4&-&- &28.3& 50.7& 66.2\\
\rowcolor[rgb]{ 0.95,  0.95,  0.95} \specialrule{0.05em}{2pt}{2pt} 
\multicolumn{11}{c}{\vspace{2pt}\textit{\textbf{Weakly-supervised instance segmentation models.}}\vspace{2pt}} \\

IRNet~\cite{DBLP:IRNet} &$\mathcal{I}$& ResNet-50&1x&Mask R-CNN&6.1 &11.7   &5.5 &-&-&-\\
BESTIE$^{\dagger}$~\cite{DBLP:BESTIE}  &$\mathcal{I}$  &  HRNet-48 &1x&Mask R-CNN& 14.3  & 28.0  & 13.2 &-&-&-\\
BBTP~\cite{DBLP:BBTP}& $\mathcal{B}$& ResNet-101&1x&Mask R-CNN& 21.1&45.5   &17.2 & 11.2 & 22.0 & 29.8\\
BoxInst~\cite{DBLP:Boxinst}& $\mathcal{B}$& ResNet-50&3x&CondInst& 32.1 &55.1&32.4 & 15.6 &34.3&43.5\\
DiscoBox~\cite{DBLP:discobox}& $\mathcal{B}$& ResNet-50&3x&SOLOv2& 32.0 &53.6&32.6 & 11.7 & 33.7 & 48.4 \\
BoxSnake~\cite{DBLP:boxsnake}& $\mathcal{B}$& ResNet-50&1x&Mask R-CNN& 31.6 &53.4&31.3 & 14.6 & 33.5 & 46.7 \\
Box2Mask~\cite{DBLP:boxlevelset}& $\mathcal{B}$& ResNet-50&1x&SOLOv2& 32.6 &55.4&33.4 & 14.7 &35.8&45.9\\
WISE-Net~\cite{DBLP:WISE-Net}&$\mathcal{P}$ &ResNet-50&1x&Mask R-CNN&7.8  &18.2  & 8.8&-&-&-\\
BESTIE$^{\dagger}$~\cite{DBLP:BESTIE}&$\mathcal{P}$& HRNet-48 &1x&Mask R-CNN&17.7 & 34.0 &16.4 &-&-&-\\

AttnShift$^{\dagger}$~\cite{DBLP:Attnshift}&$\mathcal{P}$ & ViT-B &50e&Mask R-CNN&21.2 &43.5&19.4 &-&-&-\\
DMPT~\cite{DMPT}&$\mathcal{P}$ & ViT-S &1x & Mask R-CNN&22.7 &45.5&21.5 &-&-&-\\
P2Obejct~\cite{p2object}&$\mathcal{P}$ & ResNet-50 &1x&Mask R-CNN&23.1 &43.7&22.2 &-&-&-\\
WISH~\cite{wish}&$\mathcal{P}$ & ResNet-50 &1x&Mask2Former&31.9 &-&- &-&-&-\\

\hline
\hline

 \rowcolor[rgb]{ 0.902,  0.902,  0.902} {SAM-SOLOv2}& $\mathcal{P}$& ResNet-50 &1x&SOLOv2& 24.6& 41.9  & 25.3 & 9.3& 28.6 & 38.1\\
 \rowcolor[rgb]{ 0.902,  0.902,  0.902} {MIL-SOLOv2}& $\mathcal{P}$& ResNet-50 &1x&SOLOv2& 26.8& 47.7  & 26.8 & 11.2 & 31.5 & 40.4\\
 \rowcolor[rgb]{ 0.902,  0.902,  0.902}\textbf{SAPNet (\emph{MPS})}& $\mathcal{P}$& ResNet-50 &1x&SOLOv2& \textbf{31.2}& \textbf{51.8}  & \textbf{32.3} & \textbf{12.6} & \textbf{35.1} & \textbf{47.8}\\
 \rowcolor[rgb]{ 0.902,  0.902,  0.902}\textbf{SAPNet (\emph{MPS})}& $\mathcal{P}$& ResNet-101 &3x&SOLOv2& \textbf{34.2}& \textbf{55.7}  & \textbf{36.0} & \textbf{15.6} & \textbf{39.1} & \textbf{51.1}\\
 \rowcolor[rgb]{ 0.902,  0.902,  0.902}\textbf{SAPNet}& $\mathcal{P}$& ResNet-50 &1x&SOLOv2& \textbf{30.7}& \textbf{51.2}  & \textbf{31.8} & \textbf{11.9} & \textbf{34.7} & \textbf{47.1}\\

 \rowcolor[rgb]{ 0.902,  0.902,  0.902}\textbf{SAPNet}& $\mathcal{P}$& ResNet-101&3x&SOLOv2&\textbf{34.6} &\textbf{56.0}  &\textbf{36.6} &\textbf{15.7} &\textbf{39.5}  &\textbf{52.1}  \\
 
 \rowcolor[rgb]{ 0.902,  0.902,  0.902}\textbf{SAPNet++}& $\mathcal{P}$& ResNet-50 &1x&SOLOv2& \textbf{32.4}& \textbf{53.5}  & \textbf{34.2} & \textbf{14.1} & \textbf{37.4} & \textbf{49.0}\\
 \rowcolor[rgb]{ 0.902,  0.902,  0.902}\textbf{SAPNet++}& $\mathcal{P}$& ResNet-101&3x&SOLOv2&\textbf{35.7} &\textbf{57.6}  &\textbf{37.7} &\textbf{16.4} &\textbf{40.7}  &\textbf{54.3}  \\
\specialrule{0.13em}{0pt}{0pt}
\end{tabular}}
\end{center}
\vspace{-5pt}
\caption{Mask annotation ($\mathcal{M}$), image annotation ($\mathcal{I}$), box annotation ($\mathcal{B}$) and point annotation ($\mathcal{P}$) performance on COCO-17 val. ``Ann.'' is the annotation type and ``sched.'' means schedule. $^{\dagger}$ indicates applying MRCNN refinement. The multi-scale augment training for re-training segmentation methods on 3x schedule for a fair comparison, and other experiments are on the single-scale training. SwinMR is Swin-Transformer-Mask R-CNN. SwinMR and Mask2Former use multi-scale augment strategies for SOTA. SAPNet (MPS) refers to adding MPS strategy in the appendix during training. }
\label{tab:coco_table1}
\vspace{-5pt}
\end{table*}
\subsection{Experimental Settings}
\subsubsection{Datasets and Evaluation}
\textbf{Datasets.} We conduct our experiments on four public datasets: COCO 2017~\cite{DBLP:coco}, Pascal VOC~\cite{DBLP:VOC}, Cityscapes~\cite{DBLP:cityscapes} and iSAID~\cite{DBLP:isaid}. \textbf{COCO 2017} includes 118k training and 5k validation images across 80 common categories. As the ground truth for the test set is not available, we train our model on the training set and evaluate it on the validation set.
\textbf{Pascal VOC} consists of 20 categories, we used VOC2012SBD to evaluate our methods, which contains 10,582 images for model training and 1,449 validation images for evaluation.
\textbf{Cityscapes} is a large-scale dataset designed for autonomous driving, featuring eight classes for instance segmentation. We train our model on 2,975 annotated images from the training set and validate its performance on 500 images from the validation set.
\textbf{iSAID} is a high-resolution remote sensing dataset for aerial instance segmentation, challenging due to its numerous small-scale objects in complex backgrounds. It includes 15 categories with 1,411 training and 458 validation images, totaling 655,451 instance annotations.\\
\textbf{Evaluation Metric.} For the COCO dataset, we use the mean average precision metric mAP@[.5,.95] and report a comprehensive set of performance indicators including \{$\rm AP$, $\rm AP_{50}$, $\rm AP_{75}$, $\rm AP_{small}$, $\rm AP_{middle}$, $\rm AP_{large}$\} for both MS-COCO and iSAID. For VOC12SBD, we report \(\{AP, AP_{25}, AP_{50}, AP_{75}\}\), and for Cityscapes, we present \(\{AP, AP_{50}\}\) for both detection and segmentation tasks. The $\rm mIoU_{box}$ represents the mean IoU between the predicted pseudo-boxes and ground truth boxes, and the \(mIoU_{mask}\) denotes the mean IoU between the predicted pseudo-masks and ground truth masks. These metrics assess SAPNet++'s ability to accurately select mask proposals without using the segmentation branch.
\subsubsection{Implementation Details.}
In our study, we employ the Stochastic Gradient Descent (SGD)~\cite{DBLP:SGD} optimizer. Our experiments are conducted using the mmdetection toolbox \cite{DBLP:mmdetection}, following the standard training protocols for each dataset. We use the ResNet~\cite{DBLP:resnet}, pretrained on ImageNet~\cite{DBLP:imagenet} as the backbone. For the COCO dataset, the batch size is set at four images per GPU across eight RTX4090. In contrast, the configuration utilize four GPUs for the VOC2012, Cityscapes, and iSAID datasets. Regarding the learning rates, for the COCO and VOC2012SBD datasets, the initial settings are 2$\times$\(10^{-2}\) and 2$\times$\(10^{-3}\), respectively, differing from SAPNet’s initial rate of 1.5$\times$\(10^{-2}\) for COCO. These rates are reduced by a factor of 10 at the 8th and 10th epochs. The initial learning rates, for the Cityscapes and iSAID datasets, are set at 1$\times$\(10^{-2}\) respectively.
For our SAM-top1 baseline, we supervise SOLOv2 with pseudo-masks, each being the proposal with the highest \textbf{SAM-internal confidence score} from a single point prompt. Training is conducted on RTX4090 GPUs with a batch size of 4 on per device and a learning rate of 0.01, using the standard SOLOv2 pipeline without modification.

In point distance guidance, the exponential factor \(d\) is set to 0.015. For negative proposal generator, the IoU thresholds $T_{neg1}$ and $T_{neg2}$ are 0.3 and 0.5, respectively. For box mining strategy, the $k$ is set to 3. In Cascade Local Affinity Refinement, we empirically set the Gaussian kernel size to 5 for the first two high-resolution layers and 3 for the lower-resolution layers.
In SAPNet, we select masks with scores outside of $\mathcal{M}_{srm}$ for multi-mask proposal supervision (MPS) to accelerate the convergence of segmentation from the mask bag.
SAPNet(MPS) employs the Multi-mask Proposal Supervision only during training for 12 epochs. In a 3x schedule, variations in the mask quality used for supervision can actually lead to decreased performance. Consequently, in SAPNet++, we revise this approach and no longer use Multi-mask Proposal Supervision as a method to accelerate convergence.
The training spans 12 epochs. Single-scale evaluation (1333 × 800) is used for the 1x schedule. For the 3x schedule, a multi-scale training approach is adopted, with the image's short side resized between 640 and 800 pixels (in 32-pixel increments). Inference is conducted using single-scale evaluation.
\subsection{Experimental Comparisons}
\begin{table}[tb!]
    \centering
    \resizebox{1.\linewidth}{!}{
    \begin{tabular}{l|ccccc}
    \specialrule{0.13em}{0pt}{1pt}
         Method&Ann.&Backbone&$ \rm AP_{25}$&$ \rm AP_{50}$&$ \rm AP_{75}$\\
    \specialrule{0.08em}{0pt}{0pt} 

        Mask R-CNN~\cite{DBLP:MASK-RCNN}&$\mathcal{M}$&R-50&78.0&68.8&43.3\\

        Mask R-CNN~\cite{DBLP:MASK-RCNN}&$\mathcal{M}$&R-101&79.6&70.2&45.3\\
        
        BoxInst~\cite{DBLP:Boxinst} &$\mathcal{B}$&R-50&-&59.1&34.2\\
        BoxInst~\cite{DBLP:Boxinst} &$\mathcal{B}$&R-101&-&61.4&37.0\\
        DiscoBox~\cite{DBLP:discobox} &$\mathcal{B}$&R-50&71.4& 59.8 &35.5 \\
        DiscoBox~\cite{DBLP:discobox} &$\mathcal{B}$&R-101&72.8& 62.2 &37.5 \\
        Box2Mask~\cite{DBLP:box2mask} &$\mathcal{B}$&R-50&76.9& 65.9 &38.2 \\
        Box2Mask~\cite{DBLP:box2mask} &$\mathcal{B}$&R-101&77.3& 66.6 &40.9 \\
        BESTIE~\cite{DBLP:BESTIE}& $\mathcal{I}$ &HRNet &53.5&41.7&24.2 \\
        IRNet~\cite{DBLP:IRNet} &$\mathcal{I}$&R-50&-&46.7&23.5\\
        BESTIE$^{\dagger}$~\cite{DBLP:BESTIE}& $\mathcal{I}$ &HRNet &61.2&51.0&26.6 \\
         WISE-Net~\cite{DBLP:WISE-Net}&$\mathcal{P}$&R-50&53.5&43.0&25.9 \\
         BESTIE~\cite{DBLP:BESTIE}&$\mathcal{P}$&HRNet&58.6&46.7&26.3 \\
         BESTIE$^{\dagger}$~\cite{DBLP:BESTIE}&$\mathcal{P}$&HRNet&66.4&56.1&30.2 \\
         Attnshift~\cite{DBLP:Attnshift} &$\mathcal{P}$&Vit-S&68.3 &54.4&25.4 \\
         Attnshift$^{\dagger}$~\cite{DBLP:Attnshift} &$\mathcal{P}$&Vit-S&70.3 &57.1&30.4 \\         
        P2Object~\cite{p2object}&$\mathcal{P}$&R-101&72.0&57.7&26.1 \\
       
         \hline
         \hline

         \rowcolor[rgb]{ 0.902,  0.902,  0.902}\textbf{SAPNet}&$\mathcal{P}$&R-50 &\textbf{70.3}&\textbf{60.3}&\textbf{35.4} \\ 
         \rowcolor[rgb]{ 0.902,  0.902,  0.902}\textbf{SAPNet}&$\mathcal{P}$&R-101 &\textbf{76.5}&\textbf{64.8}&\textbf{40.8} \\ 
         \rowcolor[rgb]{ 0.902,  0.902,  0.902}\textbf{SAPNet++}&$\mathcal{P}$&R-50 &\textbf{72.2}&\textbf{61.7}&\textbf{36.8} \\ 
         \rowcolor[rgb]{ 0.902,  0.902,  0.902}\textbf{SAPNet++}&$\mathcal{P}$&R-101 &\textbf{77.5}&\textbf{66.2}&\textbf{41.4} \\ 
             \specialrule{0.13em}{0pt}{0pt}
    \end{tabular}}
    \caption{Instance segmentation performance on the VOC2012 $test$ set. For a fair comparison, the segmentation network is trained on a multi-scale augment dataset using our and previous networks on Resnet101. $^{\dagger}$ indicates applying MRCNN refinement.
    }
    \label{tab:voc_12_seg}
\vspace{-10pt}
\end{table}
\subsubsection{Experiments on General Scenes}
Tab.~\ref{tab:coco_table1} compares our method with previous state-of-the-art (SOTA) approaches on COCO, spanning various supervision modalities from fully-supervised Mask ($\mathcal{M}$) to weakly-supervised Image ($\mathcal{I}$), Box ($\mathcal{B}$), and our focus, Point ($\mathcal{P}$)~\cite{DBLP:MASK-RCNN,DBLP:solov2,DBLP:condinst,DBLP:mask2former,DBLP:swin-transformer}.
In our experiments, we provide SAM with labeled points and annotations from~\cite{DBLP:cpf}. SAM then utilizes these inputs to generate subsequent mask proposals for selection and supervision. For fair comparisons, we design \textbf{two baselines}: the top-1 scored mask from SAM and traditional MIL-selected SAM mask proposals are used as SOLOv2 supervision, respectively. Tab.~\ref{tab:coco_table1} shows that our methods (SAPNet \& SAPNet++) substantially surpasses these baselines in performance.

\textbf{Comparison with point-annotated methods.} 
Our approaches achieve state-of-the-art performance. With a ResNet-50 backbone (1x schedule), SAPNet++ (32.4~$\rm AP$) surpasses previous non-SAM-based SOTAs like BESTIE~\cite{DBLP:BESTIE} and AttnShift~\cite{DBLP:Attnshift}, as well as recent SAM-based methods like DMPT~\cite{DMPT} (22.7~$\rm AP$) and WISH~\cite{wish} (31.9~$\rm AP$). This highlights the effectiveness of our spatial awareness and affinity refinement modules in pushing the performance boundary.
The performance advantage is further amplified with a longer training schedule. Under a 3x schedule with a ResNet-101 backbone, our SAPNet++ achieves 35.7~$\rm AP$. This widens the performance gap over AttnShift$^{\dagger}$~\cite{DBLP:Attnshift} (which relies on a large ViT-B~\cite{DBLP:VIT} model) to a remarkable 14.5~$\rm AP$, demonstrating our method's strong scalability and its superiority even when compared to methods that leverage large-scale models and extended training.
Importantly, our method is trained end-to-end without post-processing, achieving robust SOTA performance in point-annotated instance segmentation across multiple settings.

\textbf{Comparison with methods based on other annotation types.}  

\begin{figure*}[ht]
\begin{center}
    \includegraphics[width=0.9\linewidth, ]{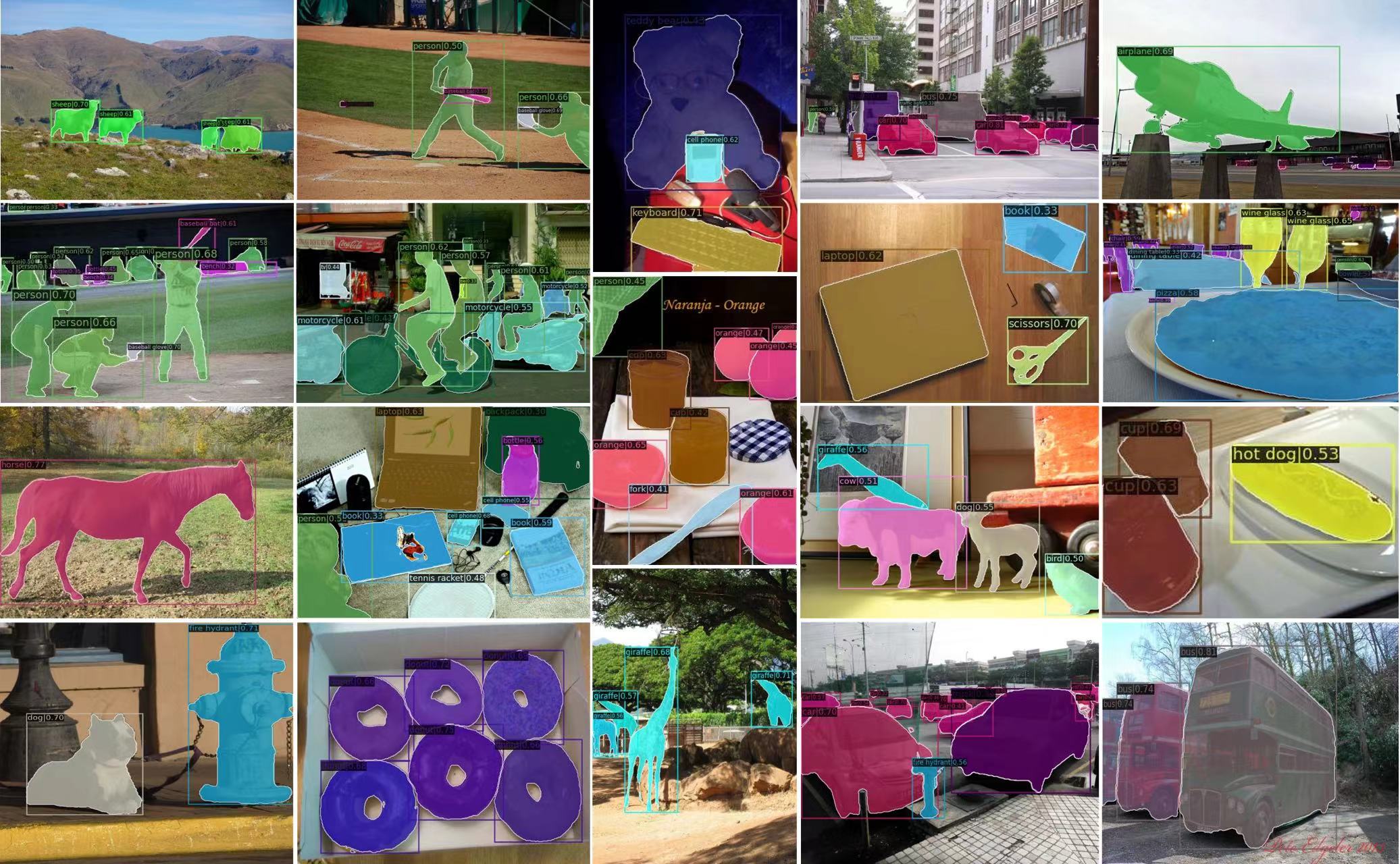}
    \captionsetup{width=0.75\linewidth}
    \caption{Visualization of instance segmentation results on the COCO train2017 dataset.
   }
\label{fig:visual_our}
\end{center}
\vspace{-15pt}
\end{figure*}

Our SAPNet has significantly elevated point annotation, regardless of its limitations in annotation time and quality compared to the box annotation. Utilizing a ResNet-101 backbone and a 3x training schedule, SAPNet surpasses the most box-annotated instance segmentation methods, achieving a 1.4 $\rm AP$ improvement over BoxInst~\cite{DBLP:Boxinst}. Even when trained on a 1x schedule with an R-50 backbone, the improved SAPNet++ surpasses the most box-supervised methods, exceeding DiscoBox~\cite{DBLP:discobox} by 0.4 \(\rm AP\), BoxInst by 0.3 \(\rm AP\), and is only 0.2 \(\rm AP\) behind the SOTA (Box2Mask) in box-supervised instance segmentation~\cite{DBLP:box2mask}. With an R101 backbone on a 3x schedule, SAPNet++ outperforms the current leading box-supervised methods. \textcolor{black}{Moreover, SAPNet++'s segmentation performance nearly matches the mask-annotated methods (92.2\% Mask2Former~\cite{DBLP:mask2former}, 86.9\% MaskDINO~\cite{DBLP:maskdino}), effectively bridging the gap between point-annotated and these techniques.}
Fig.~\ref{fig:visual_our} shows SAPNet++'s instance segmentation results on the COCO dataset, demonstrating superior segmentation performance for individual large objects and denser scenes. Our method exhibits outstanding capabilities in segmenting singular large targets and operating effectively in complex environments.

\textbf{Segmentation performance on VOC2012SBD.} Tab.~\ref{tab:voc_12_seg} compares segmentation methods under different supervisions on the VOC2012 dataset. SAPNet reports an enhancement of 7.7 $\rm AP$ over the AttnShift approach, evidencing a notable advancement in performance. Thereby, SAPNet significantly outstrips the image-level supervised segmentation methods. Additionally, SAPNet surpasses the box-annotated segmentation methods, such as BoxInst by 3.4 $\rm AP_{50}$ and DiscoBox by 2.6 $\rm AP_{50}$. 
Furthermore, the improved SAPNet++ nearly matches the performance of the SOTA box-supervised method Box2Mask, \cite{DBLP:box2mask}, achieving SOTA performance for point-prompted methods on the VOC2012 dataset with 94.3\% of the performance of Mask-RCNN.
\begin{table}[tb!]
    \centering
    \resizebox{1.\linewidth}{!}{
    \begin{tabular}{l|cc|c|c}
    \specialrule{0.13em}{0pt}{1pt}
         Method&\makebox[0.025\textwidth][c]{Ann.}&\makebox[0.02\textwidth][c]{BB.}&\makebox[0.125\textwidth][c]{COCO17 ($\rm AP$)}&\makebox[0.11\textwidth][c]{VOC12 ($\rm AP_{50}$)}\\
             \specialrule{0.08em}{0pt}{0pt}
        FPN~\cite{DBLP:FPN} &$\mathcal{B}$&R-50&37.4&75.3\\
        CASD~\cite{DBLP:CASD} &$\mathcal{I}$&VGG-16&12.8&53.6\\
        CASD~\cite{DBLP:CASD} &$\mathcal{I}$&R-50&13.9&56.8\\
        OD-WSCL~\cite{DBLP:OD-WSCL} & $\mathcal{I}$&VGG-16&13.6&56.2\\
        OD-WSCL~\cite{DBLP:OD-WSCL} & $\mathcal{I}$&R-101&14.4&-\\
        UFO$^2$~\cite{DBLP:ufo2}&$\mathcal{P}$&VGG-16&13.0&41.0\\
        UFO$^{2{\ddagger}}$~\cite{DBLP:ufo2}&$\mathcal{P}$&R-50&13.2&38.6 \\
        P2BNet-FR~\cite{DBLP:cpf}&$\mathcal{P}$&R-50&22.1&48.3\\

\hline
\hline
        \rowcolor[rgb]{ 0.902,  0.902,  0.902}\textbf{SAPNet}&$\mathcal{P}$&R-50& \textbf{32.5} &\textbf{64.8}\\
        \rowcolor[rgb]{ 0.902,  0.902,  0.902}\textbf{SAPNet}&$\mathcal{P}$&R-101& \textbf{37.2} &\textbf{68.5}\\
        \rowcolor[rgb]{ 0.902,  0.902,  0.902}\textbf{SAPNet++}&$\mathcal{P}$&R-50& \textbf{34.2} &\textbf{66.7}\\
        \rowcolor[rgb]{ 0.902,  0.902,  0.902}\textbf{SAPNet++}&$\mathcal{P}$&R-101& \textbf{38.9} &\textbf{70.2}\\

        \specialrule{0.13em}{0pt}{0pt}
    \end{tabular}}
    \caption{Object detection performance on the COCO 2017 and VOC 2012 val sets.}
    \label{tab:COCO_VOC_DET}
    \vspace{-10pt}
\end{table}
\begin{table}[tb!]
    \centering
    \resizebox{1.\linewidth}{!}{
    \begin{tabular}{l|cc|cc|cc}
    \specialrule{0.13em}{0pt}{0pt}
        \multirow{2}{*}{Method} &
        \multirow{2}{*}{\makebox[0.01\textwidth][c]{Ann.}}&  \multirow{2}{*}{Pretrain}& \multicolumn{2}{c|}{Detection} & \multicolumn{2}{c}{Segmentation}\\
         &&&$\rm AP$&$\rm AP_{50}$&$\rm AP$&$\rm AP_{50}$\\
    \specialrule{0.08em}{0pt}{0pt} 
        
        FPN~\cite{DBLP:FPN} &$\mathcal{B}$&ImageNet &37.3 &65.4&-&- \\
        FPN~\cite{DBLP:FPN} &$\mathcal{B}$&COCO &40.3 &65.3 &-&-  \\
        Mask R-CNN~\cite{DBLP:MASK-RCNN} &$\mathcal{M}$&ImageNet &39.4 &66.0&34.2&61.0 \\
        Mask R-CNN~\cite{DBLP:MASK-RCNN} &$\mathcal{M}$&COCO & 40.9&66.2&36.4&61.8\\
        CondInst~\cite{DBLP:condinst} &$\mathcal{M}$&ImageNet &- &-&33.1&58.1 \\
        BoundaryFormer~\cite{DBLP:BoundaryFormer} &$\mathcal{M}$&ImageNet &- &-&34.7&60.8 \\
        SOLOv2~\cite{DBLP:solov2} &$\mathcal{M}$&ImageNet &- &-&33.7&58.8 \\
        SOLOv2~\cite{DBLP:solov2} &$\mathcal{M}$&COCO &- &-&36.1&60.5 \\
        BoxInst~\cite{DBLP:Boxinst} &$\mathcal{B}$&ImageNet &- &-&22.4&49.0 \\
        AsyInst~\cite{DBLP:asyinst} &$\mathcal{B}$&ImageNet &- &-&24.7&53.0 \\
        BoxSnake~\cite{DBLP:boxsnake} &$\mathcal{B}$&ImageNet &- &-&26.3&54.2 \\
         UFO$^{2{\ddagger}}$~\cite{DBLP:ufo2} &$\mathcal{P}$ &ImageNet&6.6&16.4&-&-\\

        UFO$^{2{\ddagger}}$~\cite{DBLP:ufo2} &$\mathcal{P}$ &COCO&7.3&19.3&-&-\\
        P2Object~\cite{p2object} &$\mathcal{P}$&ImageNet &24.6 & 58.2& 18.6& 45.1 \\
        P2Object ~\cite{p2object} &$\mathcal{P}$&COCO &27.0 &58.2 &21.0&46.4 \\
\hline  

\hline
         \rowcolor[rgb]{ 0.902,  0.902,  0.902}\textbf{SAPNet}&$\mathcal{P}$&ImageNet &\textbf{30.3}&\textbf{56.5}&\textbf{28.5}&\textbf{53.7} \\
         \rowcolor[rgb]{ 0.902,  0.902,  0.902}\textbf{SAPNet}&$\mathcal{P}$&COCO &\textbf{32.6}&\textbf{60.2}&\textbf{30.7}&\textbf{55.3}\\
         \rowcolor[rgb]{ 0.902,  0.902,  0.902}\textbf{SAPNet++}&$\mathcal{P}$&ImageNet &\textbf{31.5}&\textbf{58.8}&\textbf{29.7}&\textbf{54.6} \\
         \rowcolor[rgb]{ 0.902,  0.902,  0.902}\textbf{SAPNet++}&$\mathcal{P}$&COCO &\textbf{33.5}&\textbf{61.3}&\textbf{32.0}&\textbf{56.7} \\
    \specialrule{0.13em}{0pt}{0pt}        
    \end{tabular}}
    \caption{Object detection and instance segmentation Performance on the Cityscape $val$ set. The ``pretrain'' means models are pretrained on ImageNet or COCO dataset.}
    \label{tab:cityscape_OD}
\vspace{-15pt}
\end{table}

\textbf{Detection performance on COCO and VOC.} As is shown in Tab.~\ref{tab:COCO_VOC_DET}, we conduct an extensive comparison using the COCO and VOC datasets. Our methodology outperforms various detection methods, including fully, image-level, and point annotations. On the COCO dataset, our method significantly improves the current SOTA P2BNet \cite{DBLP:cpf}, with an increase of 10.4 $\rm AP$ (32.5 $\rm AP$ \emph{vs.} 22.1 $\rm AP$). Under a 3x training schedule, SAPNet's detection efficacy matches that of the fully-annotated FPN \cite{DBLP:FPN}. With 3x training schedule, on the VOC dataset, our approach surpasses the previous SOTA by 8.0 $\rm AP_{50}$, achieving approximately 91\% of the fully-annotated FPN's performance. Image-level methods significantly underperform point-annotated methods on the COCO dataset, achieving only 36\% of the fully-annotated methods' performance. That highlights the advantage of point-prompted methods, optimizing the trade-off between annotation effort and detection performance.
\subsubsection{Experiments on Remoting Scenes}
To explore the effectiveness of our method in aerial scene imagery, we conduct experiments on the challenging iSAID dataset. Tab.\ref{tab:isaid} reports the results for mask AP. We compare our method with fully supervised and box-supervised methods. Compared to the fully supervised approaches, our SAPNet with an R50 backbone surpasses SOLO \cite{DBLP:SOLO}, and the enhanced SAPNet++ performs even better. It approaches the performance of the strongly supervised SOLOv2, achieving 94.5\% of its performance with an R101 backbone. Both SAPNet and SAPNet++ exceed all the current box-supervised methods with R50 backbone. While SAPNet slightly underperforms compared to Box2Mask \cite{DBLP:box2mask} with R101 backbone, SAPNet++ surpasses it by 0.7 points, achieving SOTA performance on the iSAID dataset.

\subsubsection{Experiments on Autonomous Driving Scenes}
In autonomous driving scenarios, instance segmentation is greatly challenged by the complexity of street scenes and high class imbalance (frequent appearance of cars and pedestrians, but infrequent occurrences of other categories). To further validate the generalizability of our method, we conducted experiments on the challenging Cityscapes dataset, and Tab.~\ref{tab:cityscape_OD} reports the results for object detection and instance segmentation. In object detection, SAPNet showed significant improvement over the weakly supervised method UFO$^2$~\cite{DBLP:ufo2}, achieving over 80\% of the performance of fully supervised methods FPN~\cite{DBLP:FPN} either using a ResNet-50 pretrained on ImageNet or a model pretrained on COCO. The enhanced SAPNet++ performs even better, reaching 91.7\% of the performance of Mask-RCNN \cite{DBLP:MASK-RCNN} ($\rm AP_{50}$: 56.7 \emph{vs.} 61.8), Therefore, it sets a new SOTA for point-prompted instance segmentation methods.

\begin{table}[tb!]
    \centering

	\resizebox{1.\linewidth}{!}{
	\setlength{\tabcolsep}{1.0mm}{
		\begin{tabular}{l|cccccccc}
    \specialrule{0.13em}{0pt}{1pt}
			Method & Ann.& BB. & AP &AP$_{50}$ & AP$_{75}$ & AP$_{S}$ & AP$_{M}$ & AP$_{L}$ \\
	\specialrule{0.08em}{0pt}{1pt}

			Mask R-CNN~\cite{DBLP:MASK-RCNN}  & $\mathcal{M}$ &R-50  & 34.2 & 57.5 & 36.2& 19.6& 41.4 & 47.9\\
			PolarMask~\cite{DBLP:polarmask} & $\mathcal{M}$&R-50  &27.2 & 48.5 & 27.3 & - & -& -\\
			CondInst~\cite{DBLP:condinst} &$\mathcal{M}$& R-50   & 31.8  & 56.4 & 31.5 & 15.3 & 41.0 & 50.2 \\
			SOLO~\cite{DBLP:SOLO}  & $\mathcal{M}$& R-50  & 23.5 & 43.1 & 22.6 & 7.4 & 30.5 & 43.3   \\
			SOLOv2~\cite{DBLP:solov2} & $\mathcal{M}$ & R-50   & 28.9 & 50.6 & 28.8 & 11.6 & 37.8 & 48.3  \\
			SOLOv2~\cite{DBLP:solov2} & $\mathcal{M}$ & R-101    & 32.6 &54.4 &33.4 & 13.6 & 42.4 & 54.6    \\

			BoxInst~\cite{DBLP:Boxinst} & $\mathcal{B}$ & R-50  & 17.5 & 42.9 & 11.4 & 8.8 & 23.9 & 37.2  \\
			BoxInst~\cite{DBLP:Boxinst} & $\mathcal{B}$ & R-101   & 19.1 & 45.2& 13.1 & 8.5 & 25.6 & 41.2  \\
			DiscoBox~\cite{DBLP:discobox} & $\mathcal{B}$& R-50      & 21.4 & 44.1 & 17.8 &  8.6 & 26.8 & 35.3  \\
			DiscoBox~\cite{DBLP:discobox} & $\mathcal{B}$& R-101    & 22.6& 45.3 & 19.4 & 9.2 & 28.6 & 38.7  \\
			Box2Mask~\cite{DBLP:box2mask} & $\mathcal{B}$ & R-50  & 24.3 & 48.1 & 20.7 & 9.9 & 29.9 & 40.9  \\

			Box2Mask~\cite{DBLP:box2mask} & $\mathcal{B}$& R-101   & 26.6 & 50.6 & 23.8 &10.6 & 33.6 &  47.4 \\

			\hline
			\hline

			\rowcolor[rgb]{ 0.902,  0.902,  0.902}\textbf{SAPNet}~\cite{DBLP:SAPNet} & $\mathcal{P}$ & R-50  & \textbf{24.6} & \textbf{48.7} & \textbf{21.0} & \textbf{10.1} & \textbf{30.3} & \textbf{41.4}  \\
			\rowcolor[rgb]{ 0.902,  0.902,  0.902}\textbf{SAPNet}~\cite{DBLP:SAPNet} &$\mathcal{P}$&  R-101  & \textbf{26.5} & \textbf{49.8} & \textbf{25.0} & \textbf{10.6} & \textbf{33.5} & \textbf{46.9}  \\
            \rowcolor[rgb]{ 0.902,  0.902,  0.902}\textbf{SAPNet++}~\cite{DBLP:SAPNet} & $\mathcal{P}$&  R-50  & \textbf{25.7} & \textbf{48.0} & \textbf{24.4} & \textbf{10.3} & \textbf{32.6} & \textbf{43.5}  \\
            \rowcolor[rgb]{ 0.902,  0.902,  0.902}\textbf{SAPNet++}~\cite{DBLP:SAPNet} & $\mathcal{P}$&  R-101  & \textbf{27.3} & \textbf{50.3} & \textbf{26.1} & \textbf{11.0} & \textbf{34.7} & \textbf{47.5}  \\
    \specialrule{0.13em}{0pt}{0pt}
	\end{tabular}}
	}
	\caption{Instance segmentation results on remote sensing image dataset iSAID $val$. The input size is 800$\times$800 for all models with 1x training schedule.}
	\vspace{-8pt}
	\label{tab:isaid}
\end{table}
\subsection{Ablation Studies of SAPNet++}\label{sec:ablation study}
The ablation studies of SAPNet are included in the conference version~\cite{DBLP:SAPNet} and the appendix. More ablation experiments have been conducted on COCO to further analyze SAPNet++'s effectiveness and robustness.
\subsubsection{Training Stage in SAPNet and SAPNet++}
 
\begin{figure}[!t]
  \centering
    \includegraphics[height=0.335\linewidth]{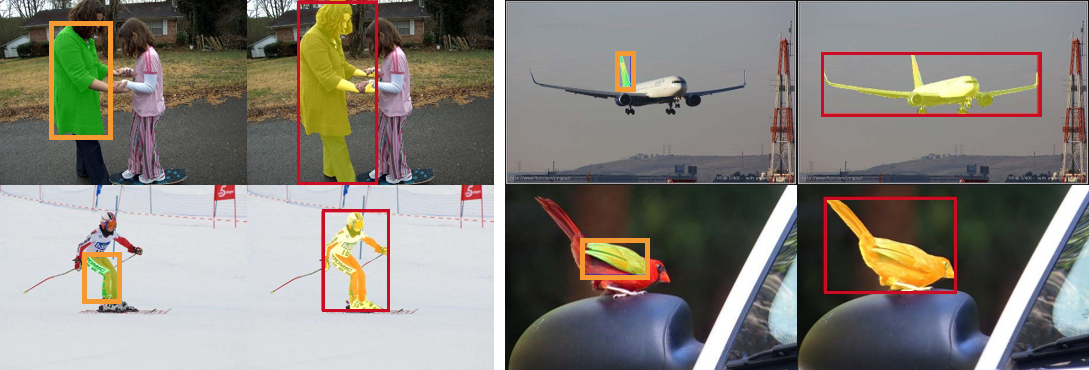}
    \caption{The comparative visualization between SAM-top1 and SAPNet++ is presented, showcasing SAM's segmentation outcomes in green masks and our results in yellow. The orange and red bounding boxes highlight the respective mask boundaries.
    }
\label{fig:sam-our}
\vspace{-5pt}
\end{figure}

The ablation study of the training stage is presented in Tab.~\ref{tab: training stage}. 
We trained SOLOv2 using the top-1 scored mask provided by SAM and compared it to the two training strategies of SAPNet. In the two-stage approach, the segmentation branch and multiple-mask supervision of SAPNet are removed. Instead, we use the selected mask to train a standalone instance segmentation model, as described by \cite{DBLP:solov2}. The end-to-end training method corresponds to the architecture illustrated in Fig.~\ref{fig:framework}. Our findings indicate that our method is more competitive than directly employing SAM (31.2 $\rm AP$ \emph{vs.} 24.6 $\rm AP$), as is shown in Fig.~\ref{fig:sam-our}. Additionally, the end-to-end training strategy boasts a more elegant model structure and outperforms the two-stage approach in overall efficiency (31.2 $\rm AP$ \emph{vs.} 30.18 $\rm AP$). Moreover, SAPNet++ shows a performance improvement of 1.7 \(\rm AP\) over the original SAPNet after enhancing spatial awareness and conducting affinity refinement.

\begin{table}[t!]
\begin{center}
\begin{tabular}{ccccccc|c}
\specialrule{0.13em}{0pt}{1pt}
PS & PDG & SR & PNPG & BMS &SASD & MLAR&$\rm mAP$ \\
\specialrule{0.08em}{0pt}{1pt}
 \checkmark&            &            &            &  & & &26.8  \\
 \checkmark& \checkmark &            &            &  & & &27.5  \\
 \checkmark& \checkmark & \checkmark &            &  & & &27.7 \\
 \checkmark& \checkmark & \checkmark & \checkmark &  & & &29.6 \\
 \checkmark& \checkmark & \checkmark & \checkmark &\checkmark& & &30.7 \\
 \checkmark& \checkmark & \checkmark & \checkmark &\checkmark &\checkmark & &31.3 \\
 \checkmark& \checkmark & \checkmark & \checkmark &\checkmark & &\checkmark &31.9 \\
 \rowcolor[rgb]{ 0.902,  0.902,  0.902}\checkmark& \checkmark & \checkmark & \checkmark &\checkmark &\checkmark &\checkmark &\textbf{32.4} \\
\specialrule{0.13em}{0pt}{0pt}
\end{tabular}
\end{center}
\caption{The effect of each component in SAPNet++: proposal selection (PS), point distance guidance (PDG), positive, negative proposals generator (PNPG), selection refinement (SR), box mining strategy (BMS), Spatial-aware Self-distillation (SASD) and Multi-level Affinity Refinement (MLAR).}
\label{tab: each component}
\vspace{-5pt}
\end{table}
\begin{table}[t!]
\begin{center}
    \resizebox{0.48\textwidth}{!}{
\begin{tabular}{l|c|c|ccc}
\specialrule{0.13em}{0pt}{1pt}
train stage on coco & type &sched. & \(\rm AP\) & \(\rm AP_{50}\)& \(\rm AP_{75}\)                \\
\specialrule{0.08em}{0pt}{0pt}

SAM-top1  & two stage &1x & 24.6 &  41.9 & 25.3 \\

SAPNet+SOLOv2 &two stage  &1x & 30.2 &   49.8   & 31.5 \\

SAPNet (MPS) & end to end &1x  & 31.2 & 51.8 & 32.3 \\
SAPNet & end to end &1x  & 30.7 & 51.2 & 31.8 \\
\rowcolor[rgb]{ 0.902,  0.902,  0.902}SAPNet++ & end to end &1x    & \textbf{32.4} & \textbf{53.5} & \textbf{34.2} \\
\specialrule{0.13em}{0pt}{0pt}
\end{tabular}}
\end{center}
\caption{The experimental comparisons of segmenters in COCO dataset, SAM-top1 is the highest scoring mask generated by SAM.}
\label{tab: training stage}
\vspace{-5pt}
\end{table}
\subsubsection{Effect of Spatial-aware Self-distillation}
As is depicted in Tab.~\ref{tab: each component}, we enhanced the original SAPNet approach by incorporating spatial-aware self-distillation (SASD). Tab.~\ref{tab: sasd} presents an ablation study of two forms of SASD: applied in the first selection branch (SASD I) and in the second refinement branch (SASD II). The results demonstrate that incorporating the SASD in either branch enhances overall performance, with SASD I contributing to an increase of 0.2 AP and SASD II boosting performance by 0.4 AP. Further, as is shown in Tab.~\ref{tab:gap}, our identified evaluation metric \(Gap\) indicates that SAPNet++ with the added SASD substantially improves matching capability. It better addresses the ``group issue'' and ``local issue'' compared to SAPNet, and significantly enhances the resulting pseudo label quality.
\begin{table}[tb!]
\begin{center}
\begin{tabular}{c|c|cccccc}
\specialrule{0.13em}{0pt}{1pt}
\multicolumn{2}{c|}{\bf SASD} & \multirow{2}{*}{\bf $\rm AP$} & \multirow{2}{*}{\bf $\rm AP_{50}$} & \multirow{2}{*}{$\rm AP_{75}$} &\multirow{2}{*}{\bf $\rm AP_s$} & \multirow{2}{*}{\bf $\rm AP_m$} & \multirow{2}{*}{$\rm AP_l$}\\
\cline{1-2} 
SASD-I & SASD-II &  &  &  \\
\specialrule{0.08em}{0pt}{1pt}
 &  & 30.7  & 51.2 & 31.8 & 11.9 & 34.7 & 47.1\\
\checkmark &  & 30.9  & 51.3 & 32.0 & 12.0 & 34.8 & 47.2\\
  & \checkmark &  31.1  & 51.7 & 32.6 & 12 & 35.0 & 47.3 \\
\rowcolor[rgb]{ 0.902,  0.902,  0.902}\checkmark  & \checkmark & \textbf{31.3} & \textbf{52.0} & \textbf{32.6} &\textbf{12.6} & \textbf{35.2} & \textbf{48.0}\\
\specialrule{0.13em}{0pt}{0pt}
\end{tabular}
\end{center}
\vspace{-5pt}
\caption{Meticulous ablation experiments in SASD}
\label{tab: sasd}
\vspace{-10pt}
\end{table}
\begin{table}[t]
	\centering 
		\begin{tabular}{lcccccc}
        \specialrule{0.13em}{0pt}{1pt}
			\textit{Affinity-level} &  AP &AP$_{50}$ & AP$_{75}$ & AP$_{S}$ & AP$_{M}$ & AP$_{L}$ \\
			\specialrule{0.08em}{0pt}{1pt}
			 \xmark &   30.7  & 51.2 & 31.8 & 11.9 & 34.7 & 47.1   \\
			low-level   & 31.0   & 51.4 & 32.2 & 12.1 & 35.0 & 47.2  \\
			high-level  &  31.3 & 51.9 & 32.4 & 12.7 & 35.2 & 47.9   \\
			\rowcolor[rgb]{ 0.902,  0.902,  0.902}low \& high-level &  \textbf{31.9} & \textbf{52.5} & \textbf{33.4} & \textbf{13.1} & \textbf{36.1} & \textbf{48.5}    \\
		\specialrule{0.13em}{1pt}{0pt}
	\end{tabular}
	\vspace{-5pt}
 \caption{The effectiveness of low-level pixel space and high-level semantic features on SAPNet++.}
	\label{tab:affinity_low_high}
\vspace{-5pt}
\end{table}
\begin{table}[tb!]
	\centering 
		\begin{tabular}{c|c|c|ccc}
        \specialrule{0.13em}{0pt}{1pt}
			\textit{Global affinity} &\textit{Local affinity}&\textit{Cascade}&  AP &AP$_{50}$ & AP$_{75}$  \\
			\specialrule{0.08em}{0pt}{1pt}
			\xmark  & \xmark & 0 &   30.7  & 51.2 & 31.8   \\
			\cmark& &  0 & 31.0   & 51.6 & 32.1   \\
			& \cmark& 1 &  31.2 & 51.7 & 32.3    \\
			\cmark& \cmark& 1&  31.6 & 52.1 & 33.0    \\
            \rowcolor[rgb]{ 0.902,  0.902,  0.902}\cmark& \cmark& 2&  \textbf{31.9} & \textbf{52.5} & \textbf{33.4}    \\
            \cmark& \cmark& 3&  31.7 & 52.3 & 33.2    \\
		\specialrule{0.13em}{1pt}{0pt}
	\end{tabular}
 \caption{The effectiveness of global and local affinity.}
	\label{tab:affinity_cascade}
\vspace{-10pt}
\end{table}
\begin{figure}[!t]
  \centering
    \includegraphics[width=1.\linewidth]{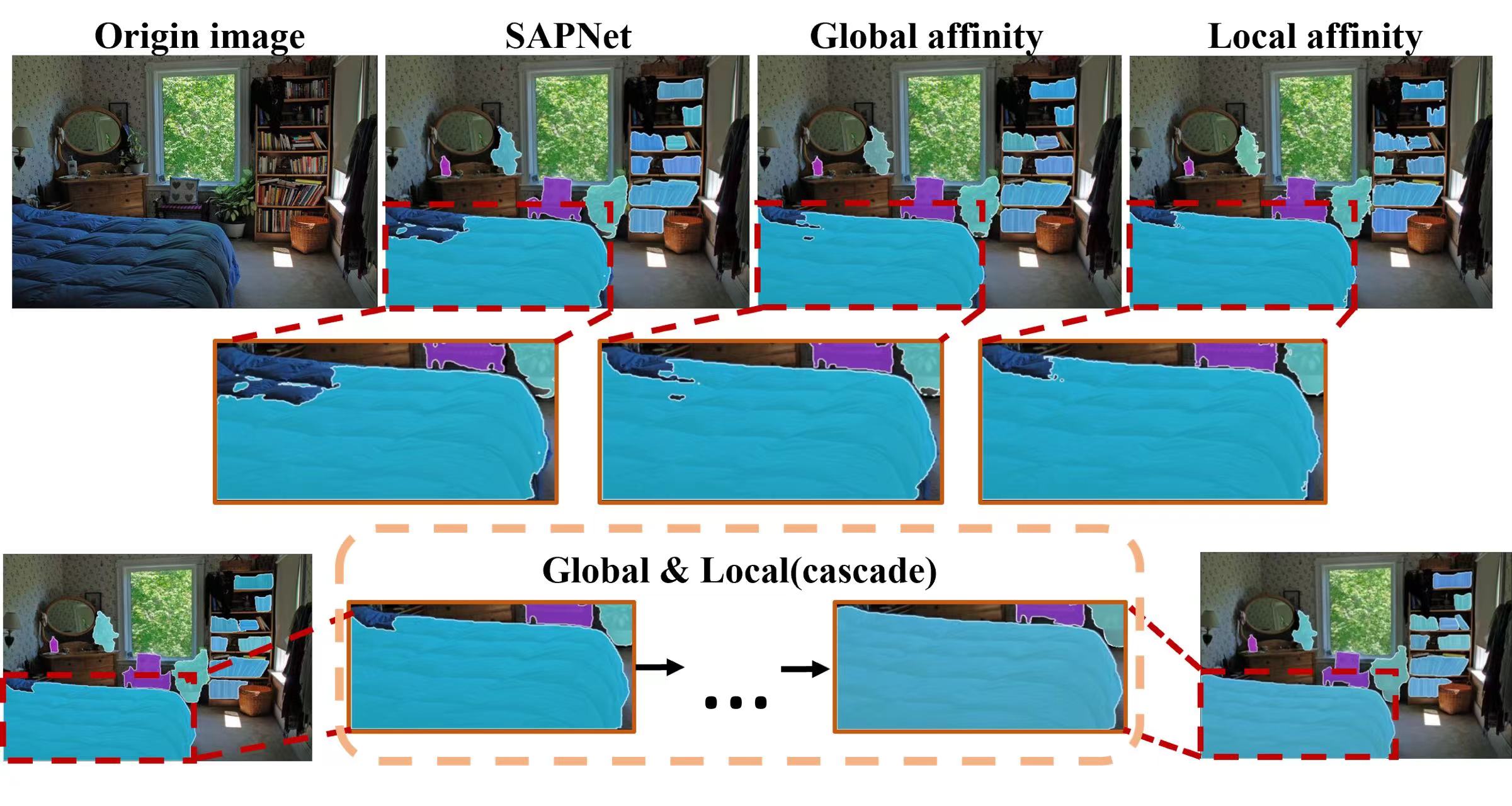}
    \vspace{-10pt}
    \caption{
    The visualization of segmentation using different affinity strategies reveals their varied impacts. with its wide propagation range, global affinity effectively manages discontinuities in large-scale masks but struggles with maintaining local consistency, often missing finer details at edges and corners. Conversely, local affinity, with its limited propagation distance, underperforms in areas where mask parts are distant. Combining both global and local affinity yields smoother segmentation masks. Using a cascading approach further improves mask completeness, striking a balance between broad coverage and detailed accuracy.
    }
\label{fig: cascade}
\vspace{-10pt}
\end{figure}
\subsubsection{Effect of Multi-level Affinity Refinement}
As is shown in Tab.~\ref{tab: each component}, using only affinity refinement on the SAPNet (without multi-mask supervision) enhances segmentation performance by 1.2 mAP (31.9 \emph{vs.} 30.7). Even when combined with SASD, affinity refinement significantly boosts segmentation results (32.4 \emph{vs.} 31.3). That indicates our affinity refinement can substantially mitigate the ``boundary uncertainty'' when supervising segmentation with incomplete masks.
Additionally, Tab.~\ref{tab:affinity_low_high} explores the use of affinity at different spatial levels (pixel-level and feature-level) through an ablation study. Employing affinity solely at the low-level pixel space allows for capturing inter-pixel color and texture relationships, resulting in a 0.3 mAP increase in segmentation performance. Utilizing affinity in the high-dimensional semantic space provides more detailed semantic information, which substantially enhances segmentation results by 0.6 mAP. Combining both low-level pixel and high-dimensional semantic space enables the segmentation network to capture both pixel-level color information and feature-level semantic details, significantly boosting segmentation performance by 1.2 mAP (31.9 \emph{vs.} 30.7).

To further validate the effectiveness of affinity refinement, we conducted ablation experiments on global and local affinity, and cascade strategies in Tab.~\ref{tab:affinity_cascade}. The experiments demonstrate that combining global and local affinity significantly enhances segmentation results. Furthermore, applying a cascade approach to local affinity further boosts segmentation performance.
Fig.~\ref{fig: cascade} illustrates the process of applying global and local affinity. Through affinity refinement, our segmentation masks progressively improve and effectively addressing the ``boundary uncertainty''. This visual representation and experimental results underscore the substantial impact of our refined affinity strategies on segmentation quality.

\subsubsection{Semantic matching Performance Analysis}\label{sec:analysis}
As presented in Tab.~\ref{tab:gap}, we conduct a statistical analysis to further validate SAPNet++'s capability to address the issue of part selection and compare the outcomes selected by the MIL with those obtained by SAPNet and SAPNet++ in the absence of segmentation branch integration. Specifically, the part problem generated by the MIL, where MIL is inclined to select proposals with a higher proportion of foreground, is exemplified in Fig. \textcolor{red}{16}. On this premise, we initially establish an evaluative criterion \( R_v = \frac{area_{mask}}{area_{box}} \), which is the ratio of the mask area to the bounding box area. Subsequently, we compute \( R_{v_i} \) for each proposal within the proposal bag corresponding to every instance across the entire COCO dataset. We select the maximum \( R_{v_{max}} \) to compute the mean value over the dataset, which is then designated as the threshold \( T_{rv} \). Ultimately, we identify the ground truth \( R_{v_{gt}} \) and objects where \( R_{v_{max}} \) exceeds \( T_{rv} \) but \(area_{box}<area_{box_{gt}}\) and calculates the discrepancy between \( R_v \) values selected by MIL, SAPNet and SAPNet++. The description is as follows:
\vspace{5pt}
\begin{equation}\footnotesize
\setlength\abovedisplayskip{2pt}
\setlength\belowdisplayskip{0pt}
\begin{aligned}
Gap_{MIL} = \sum{\frac {Rv_{MIL} - Rv_{gt}}{sum}}, \ Gap_{local} = \sum{\frac{Rv_{our} - Rv_{gt}}{sum}}.
\setlength\abovedisplayskip{0pt}
\end{aligned}
\label{Eq: area}
\end{equation} 

However, we define the metric \( Gap_{group} = \frac{labels_{bad}}{sum} \) to quantify the group issue to better demonstrate how our SAPNet++ addresses the ``group issue''. Here, \(labels_{bad}\) represents the number of pseudo labels that include annotation points from other targets within the same category, and \(sum\) denotes the total number of annotations. This formula helps to quantitatively assess the extent to which SAPNet++ mitigates the grouping of multiple targets into single pseudo labels.
Tab.~\ref{tab:gap} reveals that our proposed SAPNet has alleviated the ``local issue'' and ``group issue'' traditionally faced by MIL to some extent, and the IoU of the boxes and masks selected by SAPNet significantly exceeds that of traditional MIL approaches. Furthermore, the addition of the SASD in SAPNet++ has led to further improvements on this foundation, enhancing its capability to mitigate these issues more effectively.
\begin{table}[tb!]
\begin{center}
\begin{tabular}{l|c|c|c|c}
\specialrule{0.13em}{0pt}{1pt}
Method & $Gap_{local}$ & $Gap_{group}$ & $mIoU_{box}$ & $mIoU_{mask}$\\
\specialrule{0.08em}{0pt}{0pt}
SAM & 28.4	& 24.7	& 58.3	& 55.6 \\
MIL  & 19.9 &18.6 & 63.8 & 58.8\\
SAPNet  & 13.1 & 14.1 & 69.1 & 65.9\\
\rowcolor[rgb]{ 0.902,  0.902,  0.902}SAPNet++  & \textbf{12.4} & \textbf{13.3} & \textbf{70.4} & \textbf{67.7}\\
\specialrule{0.13em}{0pt}{0pt}
\end{tabular}
\end{center}
\vspace{-5pt}
\caption{Experimental analysis with group and local issues.}
\label{tab:gap}
\vspace{-5pt}
\end{table}

\begin{table}[tb!]
\begin{center}
    \begin{tabular}{l|cc|cc|cc}
    \specialrule{0.13em}{0pt}{1pt}
    \multirow{2}{*}{Segmenters} & \multirow{2}{*}{BB.}&\multirow{2}{*}{Sch.}& \multicolumn{2}{c|}{Detection} & \multicolumn{2}{c}{Segmentation} \\
    \cline{4-7}
                               & && $\rm AP$ & $\rm AP_{50}$ & $\rm AP$ & $\rm AP_{50}$ \\
    \specialrule{0.08em}{0pt}{1pt}
    Solov2 &R-101 &3x &38.9 &62.0 &35.7 &57.6\\
    Swin                  & Swin-S & 3x & 40.1 & 63.4 &35.8 &59.5 \\
    Mask2former                  &Swin-S&3x& 43.1 & 63.5 & 39.5 & 62.8 \\
    ConvNeXt-V2                  & ConV2-B & 3x & 46.0 & 66.8 &39.7 &63.1 \\
    \specialrule{0.13em}{0pt}{0pt}
    \end{tabular}
\caption{Advanced segmenters on COCO.}
\label{rebuttal:tab:new segmenters}
\end{center}
\vspace{-15pt}
\end{table}
\begin{table}[htbp]
\centering
\begin{tabular}{@{}lcccc@{}}
\toprule
\textbf{Type} & \textbf{Image tag} & \cellcolor{lightgray}\textbf{Point} & \textbf{Box} & \textbf{Mask} \\
\midrule
Cost (per Image)  & 1.5s/image  & \cellcolor{lightgray}1.87s & 34.5s & 239.7s \\
Cost (per Object) & --          & \cellcolor{lightgray}0.9s  & 7s   & 79.2s \\
\bottomrule
\end{tabular}
\caption{Annotation cost comparison for different supervision types.}
\label{tab:ann_cost_comparison}
\vspace{-10pt}
\end{table}
\begin{table}[]
\centering
\begin{tabular}{@{}llcl@{}}
\toprule
\textbf{Supervision Type} & \textbf{Method} & \textbf{mAP} & \textbf{\begin{tabular}[c]{@{}l@{}}Rel. Ann. Cost \\ (vs. Point/Image)\end{tabular}} \\
\midrule

\begin{tabular}[c]{@{}l@{}}Image Tag \\ (Weakest)\end{tabular} 
& BESTIE & 14.3\,mAP & 1.25$\times$ faster \\
\midrule

\textbf{\begin{tabular}[c]{@{}l@{}}Point \\ (Our Method)\end{tabular}} 
& \textbf{SAPNet++} & \textbf{32.4\,mAP} & \textbf{Base Cost} \\
\midrule

Box 
& Box2Mask & 32.6\,mAP & 18.4$\times$ slower \\
\midrule
\begin{tabular}[c]{@{}l@{}}Mask \\ (Full Supervision)\end{tabular} 
& Mask2Former & 38.7\,mAP & 128$\times$ slower \\
\bottomrule
\end{tabular}
\caption{Tradeoff Across Different Supervision Methods.}
\label{tab:performance_tradeoff}
\vspace{-10pt}
\end{table}
\subsubsection{Performance of advanced segmenters and Trade off.}
We conducted experiments with three new segmentation networks, and the results are illustrated in Tab.\ref{rebuttal:tab:new segmenters}. It is evident that integrating SAPNet++ with higher-performance segmenters substantially improves segmentation and detection performance.

\textbf{Cost-Performance Analysis.} We analyze the cost-performance trade-off to demonstrate the practical advantages of our point-supervised approach. As shown in Tab.~\ref{tab:ann_cost_comparison}, point annotation is vastly more efficient, being approximately \textbf{18.4$\times$} cheaper than bounding box annotation and \textbf{128$\times$} cheaper than full mask annotation. This significant cost reduction is achieved with minimal performance compromise. According to Tab.~\ref{tab:performance_tradeoff}, our \textbf{SAPNet++} (32.4\,mAP) achieves performance nearly identical to the strong box-supervised baseline, Box2Mask (32.6\,mAP), with only a 0.2\,mAP difference. While fully-supervised methods like Mask2Former yield higher performance (38.7\,mAP), the marginal gain comes at an impractical \textbf{128$\times$} increase in annotation cost. In conclusion, SAPNet++ establishes a superior cost-performance profile, positioning point supervision as a highly practical and scalable solution for large-scale instance segmentation dataset generation.
\section{Conclusion}
In this work, we introduce an innovative end-to-end point-prompted segmentation paradigm, SAPNet. It integrates foundational visual models with MIL to address the granularity ambiguities of point prompts and improve proposal quality. To overcome traditional MIL's ``group'' and ``local'' issues, caused by the weakness of point prompts, SAPNet employs point distance guidance and box mining strategies. Further, SAPNet++ incorporates a Spatial-aware Self-distillation, forming S-MIL, which significantly resolves these challenges. Additionally, SAPNet++ tackles the boundary uncertainty with Multi-level Affinity Refinement, enhancing segmentation performance. Our methods are validated through extensive experiments across multiple datasets.

\textbf{Acknowledge:}
This work was supported in part by the Key Deployment Program of the Chinese Academy of Sciences, China under Grant KGFZD-145-25-39, the National Natural Science Foundation of China under Grants 62272438, and Beijing Natural Science Foundation L25700.
\vspace{-5pt}

\ifCLASSOPTIONcaptionsoff
  \newpage
\fi

\small
\bibliographystyle{IEEEtran}
\bibliography{egbib}
\vspace{-15mm}
\begin{IEEEbiography}[{\includegraphics[width=1in,height=1.25in,clip]{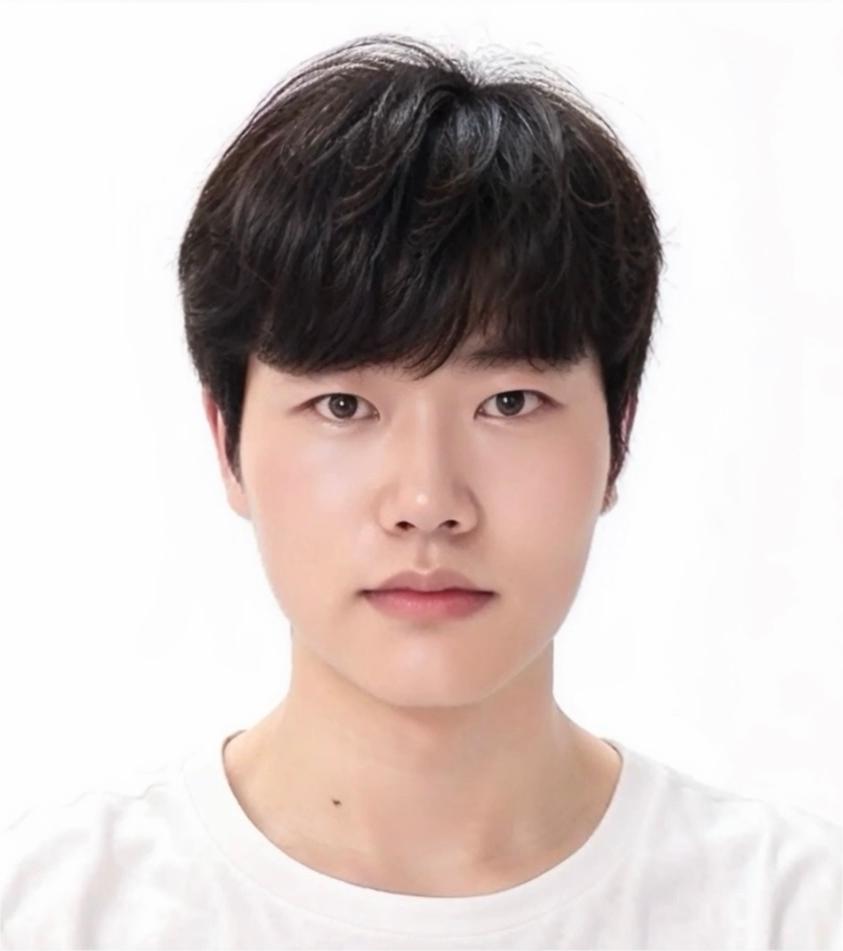}}]{Zhaoyang Wei}
received the B.E. degree in Electrical Engineering and Automation from Civil Aviation University of China, China, in 2023. He is currently pursuing the M.S. degree in Electronic and Information Engineering with the School of Electronic, Electrical, and Communication Engine, University of Chinese Academy of Sciences. His research interests include machine learning and computer vision.
\end{IEEEbiography}
\vspace{-15mm}
\begin{IEEEbiography}[{\includegraphics[width=1in,height=1.25in,clip]{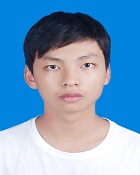}}]{Xumeng Han}
received the B.E. degree in electronic and information engineering from University of Electronic Science and Technology of China, Chengdu, China, in 2019. He is currently pursuing the Ph.D. degree in signal and information processing with the School of Electronic, Electrical, and Communication Engine, University of Chinese Academy of Sciences, Beijing, China. His research interests include machine learning and computer vision.
\end{IEEEbiography}
\vspace{-15mm}
\begin{IEEEbiography}[{\includegraphics[width=1in,height=1.25in,clip]{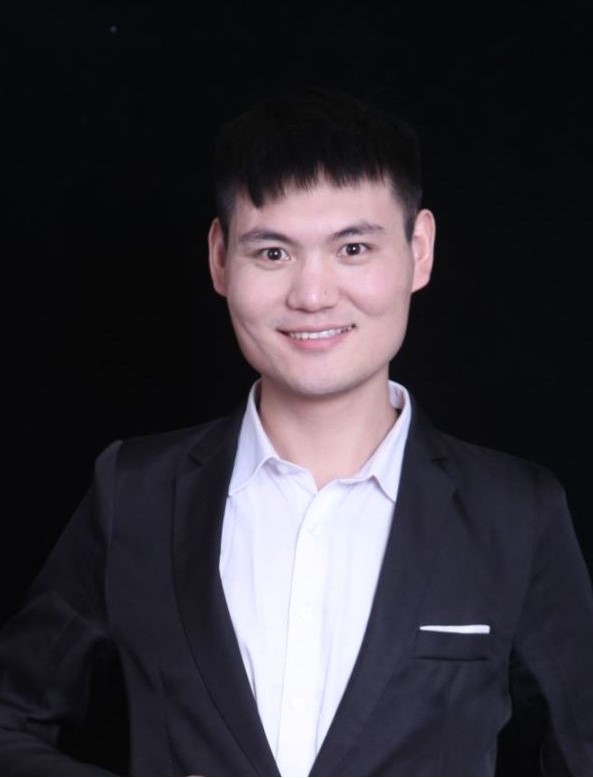}}]{Xuehui Yu}
received the B.E. degree in software engineering from Tianjin University, China, in 2017. He is currently pursuing the Ph.D. degree in signal and information processing with the School of Electronic, Electrical, and Communication Engine, University of Chinese Academy of Sciences. His research interests include machine learning and computer vision.
\end{IEEEbiography}
\vspace{-15mm}
\begin{IEEEbiography}[{\includegraphics[width=1in,height=1.25in,clip,keepaspectratio]{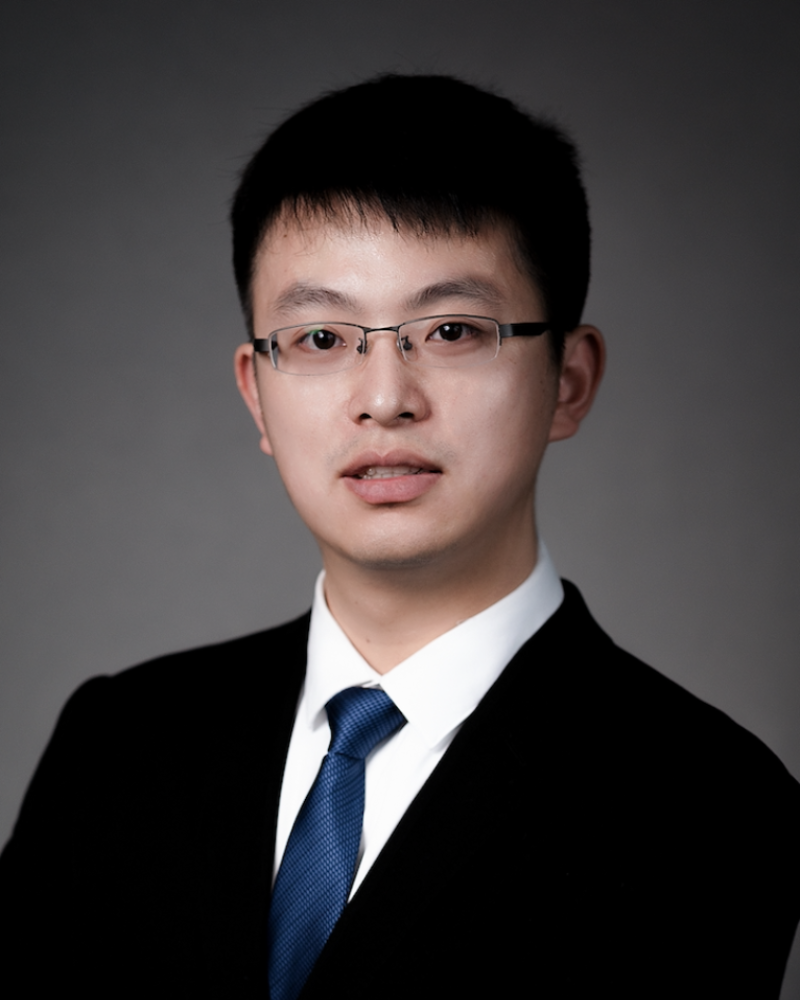}}]{Xue Yang} (Member, IEEE) received the Ph.D. in Computer Science from Shanghai Jiao Tong University, Shanghai, China, in 2023, the M.S. degree from Chinese Academy of Sciences University, Beijing, China, in 2019, and the B.E. degree from Automation, Central South University, Hunan, China, in 2016. He is currently an Assistant Professor with the School of Automation and Intelligent Sensing, Shanghai Jiao Tong University, Shanghai, China. His research interests are computer vision and machine learning. 
\end{IEEEbiography}

\vspace{-15mm}
\begin{IEEEbiography}[{\includegraphics[width=1in,height=1.25in,clip]{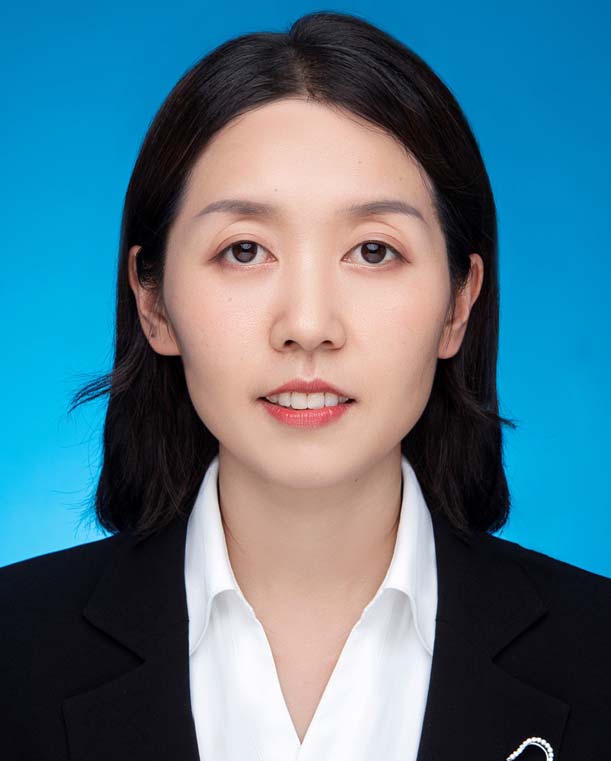}}]{Guorong Li}
(Senior Member, IEEE) received her B.S. degree in technology of computer application from Renmin University of China, in 2006 and Ph.D. degree in technology of computer application from the Graduate University of the Chinese Academy of Sciences in 2012. Now, she is an associate professor at the University of Chinese Academy of Sciences. Her research interests include object tracking, video analysis, pattern recognition, and cross-media analysis.
\end{IEEEbiography}
\vspace{-15mm}
\begin{IEEEbiography}[{\includegraphics[width=1in,height=1.25in,clip]{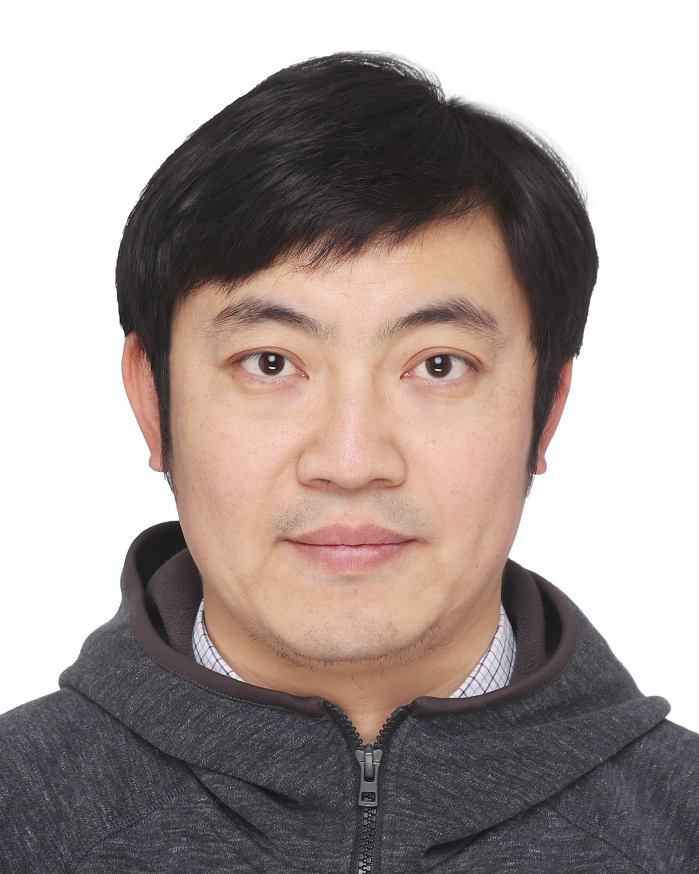}}]{Zhenjun Han}
(Member, IEEE) received the B.S. degree in software engineering from Tianjin University, Tianjin, China, in 2006 and the M.S. and Ph.D. degrees from University of Chinese Academy of Sciences, Beijing, China, in 2009 and 2012, respectively. Since 2013, he has been an Associate Professor with the School of Electronic, Electrical, and Communication Engineering, University of Chinese Academy of Sciences. His research interests include object tracking and detection.
\end{IEEEbiography}
\vspace{-15mm}

\begin{IEEEbiography}[{\includegraphics[width=1in,height=1.25in,clip]{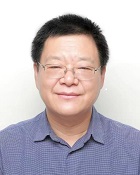}}]{Jianbin Jiao}

(Member, IEEE) received the B.S., M.S., and Ph.D. degrees in mechanical and electronic engineering from the Harbin Institute of Technology (HIT), Harbin, China, in 1989, 1992, and 1995, respectively. From 1997 to 2005, he was an Associate Professor with HIT. Since 2006, he has been a Professor with the University of Chinese Academy of Sciences, Beijing, China. His research interests include computer vision and pattern recognition.
\end{IEEEbiography}

\end{document}